\pdfoutput=1

\documentclass[11pt]{article}

\usepackage[]{EMNLP2022}

\usepackage{times}
\usepackage{latexsym}

\usepackage[T1]{fontenc}

\usepackage[utf8]{inputenc}

\usepackage{microtype}

\usepackage{multirow}
\usepackage{multicol}
\usepackage{xspace}
\usepackage{colortbl}
\usepackage{makecell}
\usepackage{graphicx}
\usepackage{booktabs}
\usepackage{amsmath}
\usepackage{arydshln}
\usepackage{amssymb}
\usepackage{tikz}
\usepackage{pgfplots}
\pgfplotsset{compat=1.8}
\usepgfplotslibrary{statistics}

\usepackage{inconsolata}

\newcommand\scone{\textsc{ScoNe}\xspace}
\newcommand\alchemy{\textsc{Alchemy}\xspace}
\newcommand\scene{\textsc{Scene}\xspace}
\newcommand\tangrams{\textsc{Tangrams}\xspace}
\newcommand\propara{\textsc{ProPara}\xspace}
\newcommand\recipes{\textsc{Recipes}\xspace}
\newcommand\ours{\textsc{LEMon}\xspace}

\newcommand{\figref}[1]{Figure~\ref{fig:#1}}

\newcommand{\tabref}[1]{Table~\ref{tab:#1}}
\newcommand{\graycell}{\rowcolor[gray]{.90}}

\newcommand{\lemon}{\raisebox{-1pt}{\includegraphics[width=1.0em]{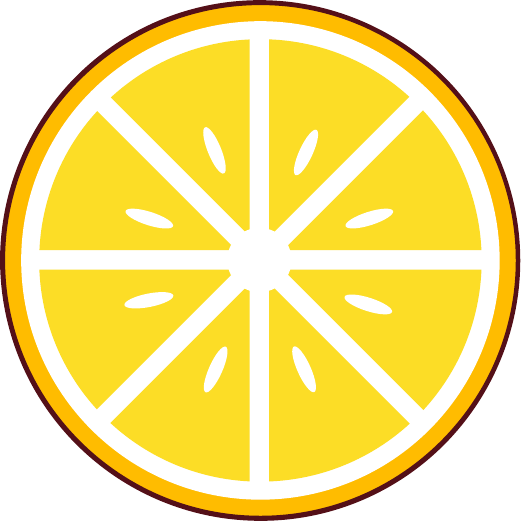}}\xspace}

\newcommand{\figinline}[1]{\raisebox{-1pt}{\includegraphics[height=1.0em]{analysis/#1}}\xspace}

\newcommand*{\affaddr}[1]{#1}
\newcommand*{\affmark}[1][*]{\textsuperscript{#1}}

\definecolor{cred}{RGB}{203,40,124}
\definecolor{cyellow}{RGB}{251,227,51}

\title{\lemon\ours: Language-Based Environment Manipulation via Execution-Guided Pre-training}

\author{Qi Shi\affmark[\textdagger]\thanks{\ \ Work done during internship at Microsoft Research Asia.}~~, Qian Liu\affmark[$\lozenge$]$^*$, Bei Chen\affmark[\S], Yu Zhang\affmark[\textdagger], Ting Liu\affmark[\textdagger], Jian-Guang Lou\affmark[\S] \\
	\affaddr{\affmark[\textdagger]Research Center for Social Computing and Information Retrieval, \\ Harbin Institute of Technology, Harbin, China} \\ \affaddr{\affmark[$\lozenge$]Beihang University, Beijing, China}; \affaddr{\affmark[\S]Microsoft Research Asia, Beijing, China} \\
	{\small \tt \{qshi, zhangyu, tliu\}@ir.hit.edu.cn}\\
	{\small \tt qian.liu@buaa.edu.cn}; {\small \tt \{beichen, jlou\}@microsoft.com}\\ 
}

\begin{document}
\maketitle
\begin{abstract}
Language-based environment manipulation requires agents to manipulate the environment following natural language instructions, which is challenging due to the huge space of the environments.
To address this challenge, various approaches have been proposed in recent work.
Although these approaches work well for their intended environments, they are difficult to generalize across environments.
In this work, we propose \ours, a general framework for language-based environment manipulation tasks.
Specifically, we first specify a task-agnostic approach for language-based environment manipulation tasks, which can deal with various environments using the same generative language model.
Then we propose an execution-guided pre-training strategy to inject prior knowledge of environments to the language model with a pure synthetic pre-training corpus.
Experimental results on tasks including \alchemy, \scene, \tangrams, \propara and \recipes demonstrate the effectiveness of \ours: it achieves new state-of-the-art results on four of the tasks, and the execution-guided pre-training strategy brings remarkable improvements on all experimental tasks\footnote{Our code is available at: \url{https://github.com/microsoft/ContextualSP}}.
\end{abstract}

\section{Introduction}
\label{intro}

\begin{figure}[t]
    \centering
    \includegraphics[width=0.98\columnwidth]{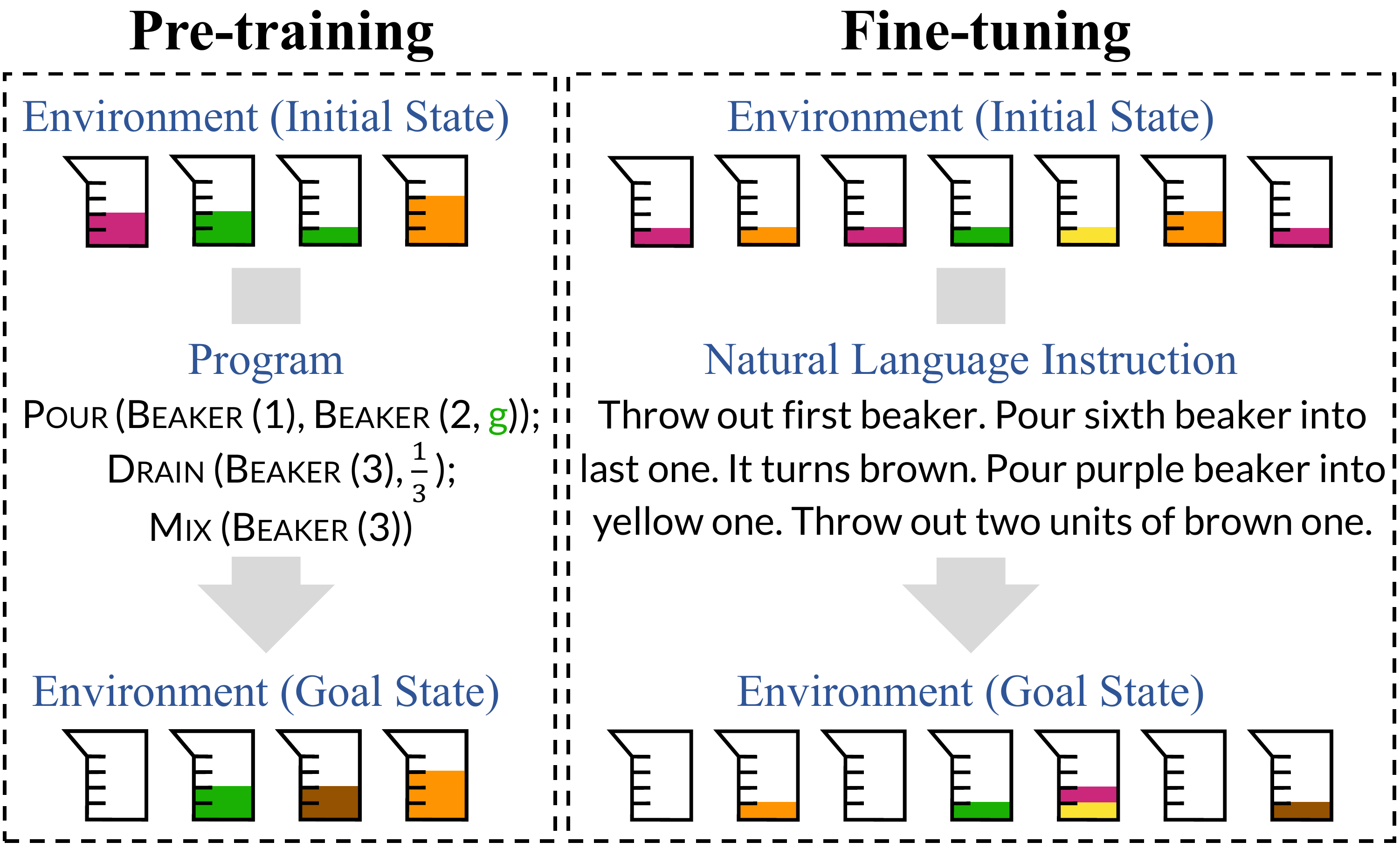}
    \caption{The schematic illustration of the pre-training (\textbf{left}) and fine-tuning (\textbf{right}) procedure of \ours. The environment is from \alchemy \protect\cite{long-etal-2016-simpler}. 
    In the pre-training stage, the input of \ours includes an initial environment state and a program, and the goal environment state is served as the supervision. The fine-tuning stage is similar to the pre-training stage, except that the program in the model input is replaced by the natural language instruction. }
    \label{fig:task_intro}
\end{figure}

Building agents that can understand human language and accordingly manipulate the environment around them has been a long-standing goal of artificial intelligence \cite{winograd1971procedures}. Various tasks focus on this scene, including collaborative building \cite{narayan-chen-etal-2019-collaborative}, state tracking \cite{dalvi-etal-2018-tracking,tandon-etal-2020-dataset} and instruction following \cite{andreas-klein-2015-alignment,long-etal-2016-simpler,suhr-etal-2019-executing}. 
What these tasks have in common is that the agents are required to manipulate the environment based on the natural language. 
To seize the commonality of existing tasks,
we define such tasks as language-based environment manipulation (LEM) tasks. 
Generally, these tasks are challenging due to the large exploration space of the environment itself and the complexity of human-agent interactions.
For example, in the environment shown in \figref{task_intro}, the agent needs to manipulate seven beakers with various colored liquids correctly according to the long instruction.

To address these challenges, recent work have proposed various specialized models to deal with different environments \cite{suhr-artzi-2018-situated,dalvi-etal-2018-tracking,gupta-durrett-2019-tracking,tang-etal-2020-understanding-procedural}.
Although these models work well, they are difficult to generalize across environments since they contain environment-specific modules.
For example, \citet{suhr-artzi-2018-situated} design different encoder modules for different environments.

Different from previous work focusing on specialized models, we argue that with formulating LEM tasks as sequence generation problems, the family of generative language models (GLMs), such as BART \cite{lewis-etal-2020-bart}, can be an environment-generic agent for various environments.
Taking advantage of GLMs, such a task-agnostic solution greatly reduces the difficulty of modeling different environments.
However, GLMs generally lack prior knowledge of downstream environments since they have not seen even similar ones during pre-training.
To unleash the power of GLMs in downstream environments, we argue that GLMs should be continually pre-trained to understand these environments, and the pre-training should engage GLMs to explore as much of the environment space as possible.
We believe if GLMs can understand the environment well, they will more easily manipulate the environment with respect to human language.

Inspired by the above, in this paper, we propose \ours (for \textbf{L}anguage-based \textbf{E}nvironment \textbf{M}anipulati\textbf{on} via Execution-guided Pre-training), a general framework for LEM tasks.
As shown in~\figref{task_intro}, \ours consists of two parts: 
1) A task-agnostic approach that uses the same protocol to tackle different LEM tasks (right). 2) An execution-guided pre-training strategy, which injects prior knowledge about environments into the GLM (left).
For the first part, we employ the popular BART \cite{lewis-etal-2020-bart} as the model backbone, and take five representative tasks \alchemy, \scene, \tangrams \cite{long-etal-2016-simpler}, \propara \cite{dalvi-etal-2018-tracking} and \recipes \cite{bosselut2017simulating} as the testbed.
For the pre-training part, it is to engage our model to explore the environment space.
Considering that the environment space mainly consists of the state space (i.e., valid environment states) and the action space (i.e., possible actions to manipulate the environment), we suggest pre-training the model via synthesizing data involving these two spaces.
Specifically, given an environment, we begin with randomly sampling its relevant initial states and programs\,\footnote{A program consists of a sequence of functions, of which each function is an action or a composition of actions.}.
With feeding the random initial state and the random program as input for \ours, we leverage the goal state after executing the program as supervision for \ours.
Since the program execution is easy to carry out in symbolic environments\,\footnote{Symbolic environments stand for the environments that can be represented by semantic symbols.}, 
our execution-guided pre-training is suitable for various symbolic environments.
Meanwhile, since the random initial states and the programs can be sampled systematically, we can readily obtain a large-scale high-quality pre-training corpus without human labeling or data cleaning.
To the best of our knowledge, \ours is the first work to explore pre-training in language-based environment manipulation.
In summary, the main contributions of our framework \ours are three-fold:
\begin{itemize}
    \item We suggest a task-agnostic approach that can be tailored to various environments. By formulating LEM tasks as sequence generation problems, our approach leverages one architecture to tackle them.
    \item We propose a novel execution-guided pre-training strategy, which can inject prior knowledge of environments by continually pre-training with only synthetic data.
    \item Experimental results on five tasks demonstrate that our task-agnostic approach is comparable or prior to previous systems, and our pre-training strategy further improves the performance by a significant margin (e.g., +$4.1$\% on \alchemy). Finally, our approach achieves new state-of-the-art results on \alchemy, \scene, \propara, and \recipes.
\end{itemize}

\begin{figure*}[t]
    \centering
    \includegraphics[width=0.85\textwidth]{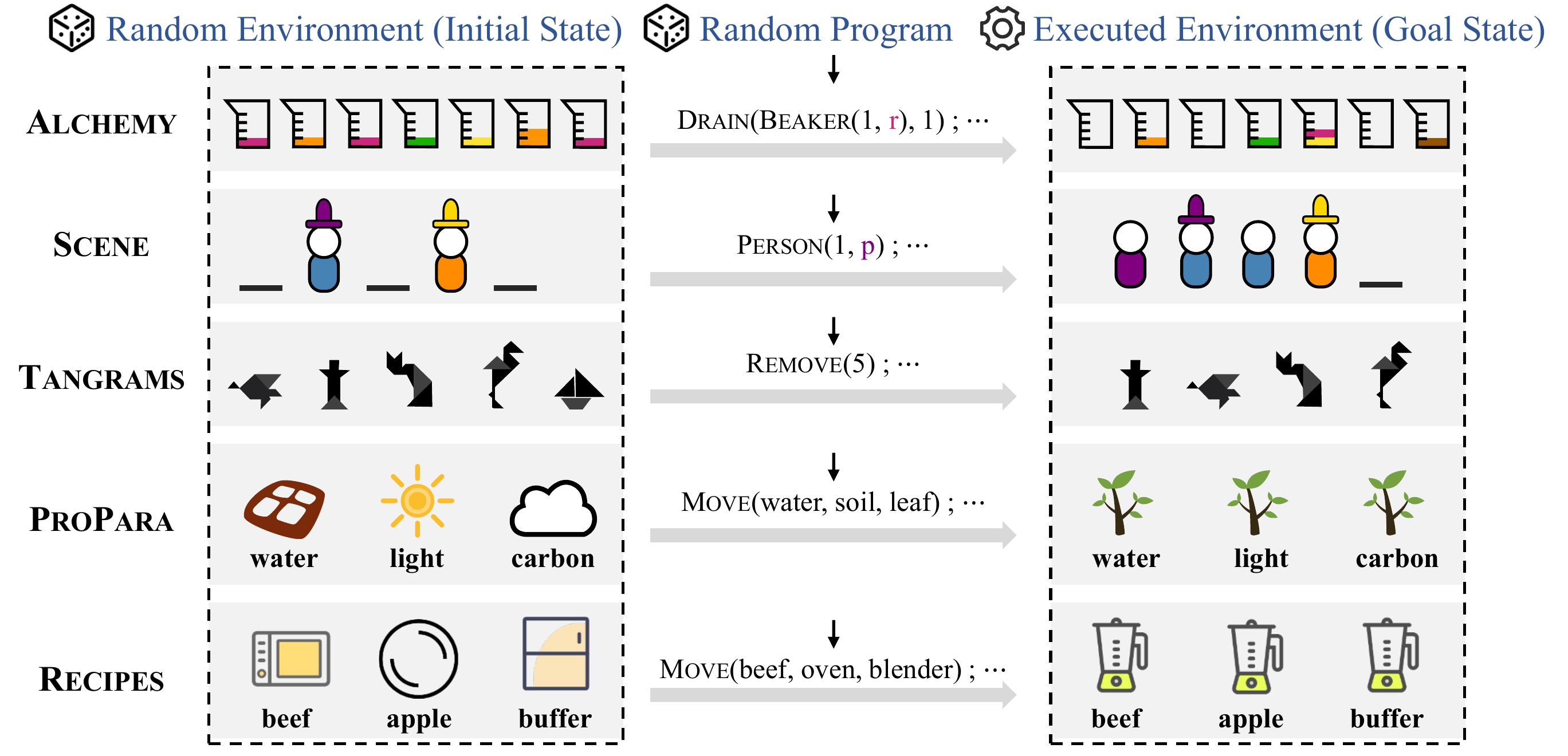}
    \caption{The illustration of the pre-training procedure of \ours framework on five tasks including \alchemy, \scene, \tangrams, \propara and \recipes.}
    \label{fig:framework}
\end{figure*}

\section{\ours Framework}
\label{sec:framework}
We now discuss the \ours framework (\figref{task_intro}) in more detail.
Specifically, we introduce the task-agnostic approach for language-based environment manipulation (\S \ref{fine-tune}) and the execution-guided pre-training (\S \ref{pre-training}).
As for \ours instantiations for different tasks, we leave the descriptions to \S \ref{instantiation}.

\subsection{Task-Agnostic Approach for LEM}
\label{fine-tune}

As mentioned in \S \ref{intro}, the existence of environment-specific modules makes previous models difficult to generalize across environments.
To eliminate this issue, we propose a task-agnostic approach to tackle different environments.

\paragraph{Task Formulation}

An environment space consists of a state space and an action space.
And a state can be further decomposed into a set of entities (e.g., beakers in \alchemy) and properties (e.g., colors in \alchemy).
Generally, the goal of LEM tasks is to manipulate the environment state with natural language.
Formally, given an initial environment state $S_0$, the goal of LEM tasks is to predict the goal environment state $S$ based on the human language instruction $I$.
In most cases, the LEM task is performed in an interactive manner, and there would be a sequence of context-dependent instructions.
Again, given an initial environment state $S_0$ and a sequence of natural language instructions $\mathbf{I}=(I_1,I_2,...,I_T)$, where $T$ stands for the total number of instructions in one conversation, the goal turns to predict the goal environment states at each step as $\mathbf{S}=(S_1,S_2,...,S_T)$.
In the following, we use the conversational formulation to illustrate.

\paragraph{Model Architecture}

With formulating LEM tasks as sequence generation problems, we leverage BART \cite{lewis-etal-2020-bart}, a powerful encoder-decoder language model, to generate the goal environment state token-by-token.
Formally, at $t$-th step, the input to our model consists of three parts, namely the initial environment state $S_0$, the history $(I_1,I_2,...,I_{t-1})$ and the current instruction $I_{t}$.
Following previous work \cite{qian2020how}, we directly concatenate the history and the current instruction to form $\mathbf{I}_{t}=(I_1,I_2,...,I_{t})$, which contains all historical instructions.
The final input to our model is the concatenation of $S_0$ and $\mathbf{I}_{t}$ with a \texttt{\small [SEP]} token as a separator between them.
The output is the corresponding goal environment state $S_t$.

\subsection{Execution-Guided Pre-training}
\label{pre-training}
We propose an execution-guided pre-training strategy to explore the environment space as much as possible through synthetic data.
In the following, we will introduce the pre-training task and the pre-training corpus generation procedure in turn.

\paragraph{Pre-training Task}

As described in \S \ref{intro}, to encourage the model to understand and explore the environment, \ours adopts the \textit{program execution} as the pre-training task.
Formally, given a randomly sampled initial environment state $S_0$ and a randomly sampled program $A$, the model is pre-trained to predict the goal environment state $S$, as shown in \figref{framework}.
Such a pre-training task fulfills our expectations of both environment exploration and environment understanding, which can be explained from two aspects.
From the input perspective, such a task involves all essential elements of an environment (i.e., state and action).
Together with large-scale random sampling, it allows the model to fully explore the environment space.
From the output perspective, such a task is challenging -- the model must understand the environment to predict $S$ correctly. 
Meanwhile, the program execution as a pre-training task is highly flexible.
As shown in \figref{framework}, it works well for five different tasks.
In the implementation of pre-training, we first concatenate $S_0$ and $A$ using the same \texttt{\small [SEP]} token as a separator and then feed the concatenated sequence into \ours.
The pre-training supervision, i.e., the goal environment state $S$, is obtained from a task-dependent executor.
In LEM tasks, the executor is designed to interpret each task-dependent program and change the current environment state to another state accordingly.
In practice, the executor can be easily implemented since the environments are symbolic.

\begin{table}[t]
    \small
    \centering
    \begin{tabular}{rl}
         $\langle$state$\rangle$ $~~\to$ & $\langle$action$\rangle$; $\langle$state$\rangle$ | $\langle$action$\rangle$ \\
        $\langle$action$\rangle$  $~~\to$ & $\langle$mix$\rangle$ | $\langle$pour$\rangle$ | $\langle$drain$\rangle$ \\
          $\langle$mix$\rangle$ $~~\to$ & \textsc{Mix} ($\langle$beaker$\rangle$) \\
          $\langle$pour$\rangle$ $~~\to$ & \textsc{Pour} ($\langle$beaker$\rangle$, $\langle$beaker$\rangle$) \\
         $\langle$drain$\rangle$ $~~\to$ & \textsc{Drain} ($\langle$beaker$\rangle$, $\langle$integer$\rangle$) ~| \\
         & \textsc{Drain} ($\langle$beaker$\rangle$, $\langle$fraction$\rangle$) \\ 
          $\langle$beaker$\rangle$ $~~\to$ & \textsc{Beaker} ($\langle$index$\rangle$) ~| \\
          & \textsc{Beaker} ($\langle$index$\rangle$,  $\langle$color$\rangle$) \\
          $\langle$index$\rangle$ $~~\to$ & $1$ | $2$ |  $\cdots$ | $7$ | $-1$ | $-2$ | $\cdots$ | $-7$ \\
         $\langle$color$\rangle$ $~~\to$ & r | g | o | p | y | b \\
         $\langle$integer$\rangle$ $~~\to$ & $1$ | $2$ | $3$ | $4$ \\
         $\langle$fraction$\rangle$ $~~\to$ & $\frac{1}{2}$ | $\frac{1}{3}$ | $\frac{1}{4}$ | $\frac{2}{3}$ | $\frac{2}{4}$ | $\frac{3}{4}$ \\
    \end{tabular}
    \caption{Grammar rules of the program used in \alchemy. The grammar rules of other domains and the descriptions can be found in Appendix \ref{all-grammar}.}
    \label{tab:alchemy_grammer_short}
\end{table}

\paragraph{Pre-training Corpus Generation}

Unlike most pre-training work that employs web crawling to collect pre-training corpus, we synthesize the pre-training corpus directly by randomly sampling the environment states and programs.
Compared to human language, high-quality environment states and programs are easier to sample since they are highly structured.
As introduced above, each pre-training example contains a sampled initial environment state, a sampled executable program, and a goal environment state obtained from the executor.
One by one, the pre-training corpus can be generated by repeating the sampling process.
Concretely, for the initial environment state sampling, it can be achieved by randomly selecting a valid value for each property defined in the corresponding environment.
As for the program sampling, a valid program can be generated by randomly selecting a valid function and then randomly sampling from all suitable parameters of the selected function.
The valid values for each property and function will be discussed later.

\begin{table*}[t]
    \centering
    \small
    \begin{tabular}{p{5.5cm}ccccccccc}
    \toprule
        \multirow{2}{*}{\textbf{Models}}
        & \multicolumn{3}{c}{\alchemy} & \multicolumn{3}{c}{\scene} & \multicolumn{3}{c}{\tangrams} \\
        \cmidrule(lr){2-4}
        \cmidrule(lr){5-7}
        \cmidrule(lr){8-10}
        & Inst & 3utts & 5utts & Inst & 3utts & 5utts & Inst & 3utts & 5utts \\
        \midrule
        \graycell \multicolumn{10}{c}{\textit{Fully Supervised Approaches}} \\
        \graycell \cite{fried-etal-2018-unified} & -- & -- & $72.0$ & -- & -- & $72.7$ & -- & -- & $69.6$ \\
        \graycell \cite{huang2018flowqa} & -- & -- & $76.4$ & -- & -- & $74.5$ & -- & -- & $72.3$ \\
        \graycell \cite{yeh2019flowdelta} & -- & -- & $76.1$ & -- & -- & $75.1$ & -- & -- & $72.5$ \\
        \multicolumn{10}{c}{\textit{Weakly Supervised Approaches}} \\
        \cite{long-etal-2016-simpler} & -- & $56.8$ & $52.3$ & -- & $23.2$ & $14.7$ & -- & $64.9$ & $27.6$ \\
        \cite{guu-etal-2017-language} & -- & $66.9$ & $52.9$ & -- & $64.8$ & $46.2$ & -- & $65.8$ & $37.1$ \\
        \cite{suhr-artzi-2018-situated} \textit{w.} \textsc{Reinforce} & $89.1$ & $74.2$ & $62.7$ & $87.1$ & $73.9$ & $62.0$ & $86.6$ & $80.8$ & $\mathbf{62.4}$ \\
         \cite{suhr-artzi-2018-situated} \textit{w.} \textsc{Heuristic} & $89.4$ & $73.3$ & $62.3$ & $88.8$ & $78.9$ & $66.4$ & $86.6$ & $81.4$ & $60.1$ \\
        \midrule
        \ours & $\mathbf{97.1}$ & $\mathbf{85.3}$ &  $\mathbf{75.4}$ & $\mathbf{92.7}$ &  $\mathbf{85.8}$ & $\mathbf{72.3}$ & $92.3$ & $82.4$ & $60.0$ \\
        ~~~~\textit{w.o.} pre-training & $96.9$ & $84.0$ & $71.3$ & $91.6$ & $83.1$ & $68.9$ & $\mathbf{92.8}$ & $\mathbf{83.4}$ & $56.7$ \\
        \bottomrule
    \end{tabular}
    \caption{Experimental results on the test set of \alchemy, \scene and \tangrams. 
    Fully supervised approaches (in grey background) are the approaches that use \textbf{annotated} programs as labels, while weakly supervised approaches are the approaches that no golden program is provided.
    Although the comparisons are \textbf{not fair}, we report the results of fully supervised approaches for reference.
    Note that our ablation \textit{w.o.} pre-training is identical to fine-tuning BART on the downstream task, and the same for \tabref{propara_result} and \tabref{recipes_result}.
    }
    \label{tab:scone_result}
\end{table*}

\section{\ours Instantiations}
\label{instantiation}
To demonstrate the capabilities of \ours, we apply our framework on five exemplary tasks, namely, \alchemy, \scene, \tangrams, \propara, and \recipes.
Examples of each task are shown in Figure \ref{fig:framework}, including visualizations of the initial environment state and the goal environment state, as well as a schematic representation of the program.
In this section, for each task, we elaborate the definition of the environment and the applied program to instantiate \ours.

\subsection{\alchemy}
\paragraph{Environment State Definition}
The environment state in \alchemy contains seven beakers, each containing up to four units of colored chemicals. 
Each environment state contains three properties, including beaker IDs (from $1$ to $7$), liquid colors (brown, green, orange, purple, red, and yellow), and liquid amounts (from $0$ to $4$).
Figure \ref{fig:framework} shows an example, and the example initial environment state can be represented as \texttt{1:r|2:o|3:r|4:g|5:y|6:oo|7:r} in text, where different letters represent different colors.
Note that if a beaker does not contain any liquid, it can be represented by \texttt{\_}. And \texttt{|} stands for the delimiter that splits the state of each beaker, which is also applicable for the following tasks. 

\paragraph{Program Definition}
The action space of \alchemy contains three kinds of actions to manipulate the environment, namely, \textsc{Pour}, \textsc{Drain} and \textsc{Mix}. 
We use the program proposed by \citet{guu-etal-2017-language}, where the functions are the same as the actions defined in the environment. 
The detail program grammar is shown in Table \ref{tab:alchemy_grammer_short}.

\subsection{\scene}
\paragraph{Environment State Definition}
The environment state in \scene contains ten positions, with up to one person in each position.
A person is defined by a shirt color and optionally a hat color.
Formally, each environment state contains three properties, including position IDs (from $1$ to $10$), shirt colors (brown, green, orange, purple, red, and yellow), and hat colors (the same as shirt colors).
As shown in Figure \ref{fig:framework}, the example initial environment state can be represented as 
\texttt{1:\_\_|2:bp|3:\_\_|4:oy|5:\_\_} 
(only five positions are shown in Figure \ref{fig:framework} for brevity) in text, where the first character in each position represents the shirt color and the second one represents the hat color. \texttt{\_} indicates either an empty position or a person without a hat. Note that the hat can only appear when the position is occupied.

\paragraph{Program Definition}
Four actions are defined in the \scene environment to manipulate the environment, namely, \textsc{Enter}, \textsc{Leave}, \textsc{Move} and \textsc{Trade-hats}.
For the program, we use the one proposed by \citet{suhr-artzi-2018-situated}. The functions include \textsc{Person}, \textsc{Hat}, \textsc{RmPerson} and \textsc{RmHat}, which represent inserting\,/\,removing a person\,/\,hat in the state.
The permutations of the defined functions in the program are sufficient to represent all actions defined in the environment.

\subsection{\tangrams}
\paragraph{Environment State Definition}
The environment state in \tangrams contains a list of up to five unique objects. 
Similarly, the environment state can be represented by the object indexes (from $1$ to $5$) and the object names (\texttt{A}, \texttt{B}, \texttt{C}, \texttt{D}, and \texttt{E}).
For example, the initial environment state in Figure \ref{fig:framework} can be represented as 
\texttt{1:A|2:B|3:C|4:D|5:E}. The same object cannot appear in one environment state. If the number of objects is less than $5$, we fill the sequence with \texttt{\_} to make it $5$ in length.

\paragraph{Program Definition}
Three actions are involved to manipulate the \tangrams environment, namely, \textsc{Add}, \textsc{Remove} and \textsc{Swap}.
And we use the program proposed by \citet{suhr-artzi-2018-situated}, which defines the functions including \textsc{Insert} and \textsc{Remove}. 
Similar to the two kinds of programs mentioned above, permuting these two functions can achieve the goal of representing all actions defined in the environment. 

\subsection{\propara \& \recipes}
\paragraph{Environment State Definition}
The \propara environment describes real-world scientific processes such as photosynthesis, erosion, etc.
Each environment state in \propara contains a set of entity participants and their corresponding locations, and the locations vary with the natural language procedural text being described.
Unlike the three environments mentioned above, the properties of an environment state in \propara are not fixed, but are dynamically constructed from the natural language text.
Figure \ref{fig:framework} shows an example, where the initial state stands for participants water, light, carbon are located in locations soil, sun, cloud respectively. 
The environment state in \propara can be naturally represented in key-value format.
For example, the initial state in Figure \ref{fig:framework} can be represented as \texttt{ent:water|light|carbon loc:soil|sun|cloud}, 
here \texttt{ent:} and \texttt{loc:} are special tokens that indicate the boundaries of entity participants and locations, respectively.

\paragraph{Program Definition}
In the \propara environment, the procedural text describes four actions, namely, \textsc{Create}, \textsc{Move}, \textsc{Destroy} and \textsc{None}.
In practice, we use the program proposed by \citet{dalvi-etal-2019-everything}, in which the functions also contain \textsc{Create}, \textsc{Move} and \textsc{Destroy}, which are aligned with the action space of \propara.
As for \recipes, the environment describes the state tracking process in the cooking domain. And the definition of the environment states and the programs are similar with \propara.

\section{Experiments}
In this section, we compare \ours with baseline methods on the tasks discussed in \S \ref{instantiation} to demonstrate its effectiveness.
Due to space limitation, we do not introduce these baselines below.

\begin{table*}[t]
    \centering
    \small
    \scalebox{0.95}{
    \begin{tabular}{p{6.4cm}cccccccc}
    \toprule
        \multirow{2}{*}{\textbf{Models}}
        & \multicolumn{5}{c}{\textbf{Sentence-Level}} & \multicolumn{3}{c}{\textbf{Document-Level}} \\
        \cmidrule(lr){2-6}
        \cmidrule(lr){7-9}
         &  {\scriptsize Cat-1} & {\scriptsize Cat-2} & {\scriptsize Cat-3} & {\scriptsize Macro-Avg} & {\scriptsize Micro-Avg} & {\scriptsize Precision} & {\scriptsize Recall} & {\scriptsize F1} \\
        \midrule
        EntNet \cite{henaff2016tracking} & $51.6$ & $18.8$ & $7.8$ & $26.1$ & $26.0$ & $54.7$ & $30.7$ & $39.4$ \\
        QRN \cite{seo2016query} & $52.4$ & $15.5$ & $10.9$ & $26.3$ & $26.5$ & $60.9$ & $31.1$ & $41.1$ \\
        ProLocal \cite{dalvi-etal-2018-tracking} & $62.7$ & $30.5$ & $10.4$ & $34.5$ & $34.0$ & $81.7$ & $36.8$ & $50.7$ \\
        ProGlobal \cite{dalvi-etal-2018-tracking} & $63.0$ & $36.4$ & $35.9$ & $45.1$ & $45.4$ & $61.7$ & $48.8$ & $51.9$ \\
        AQA \cite{ribeiro2019predicting} & $61.6$ & $40.1$ & $18.6$ & $39.4$ & $40.1$ & $62.0$ & $45.1$ & $52.3$ \\
        ProStruct \cite{tandon-etal-2018-reasoning} & -- & -- & -- & -- & -- & $74.3$ & $43.0$ & $54.5$ \\
        XPAD \cite{dalvi-etal-2019-everything} & -- & -- & -- & -- & -- & $70.5$ & $45.3$ & $55.2$ \\
        LACE \cite{du-etal-2019-consistent} & -- & -- & -- & -- & -- & $75.3$ & $45.4$ & $56.6$ \\
        KG-MRC \cite{das2018building} & $62.9$ & $40.0$ & $38.2$ & $47.0$ & $46.6$ & $69.3$ & $49.3$ & $57.6$ \\
        ProGraph \cite{zhong2020heterogeneous} & $67.8$ & $44.6$ & $41.8$ & $51.4$ & $51.5$ & $67.3$ & $55.8$ & $61.0$ \\
        IEN \cite{tang-etal-2020-understanding-procedural} & $71.8$ & $47.6$ & $40.5$ & $53.3$ & $53.0$ & $69.8$ & $56.3$ & $62.3$ \\
        NCET \cite{gupta-durrett-2019-tracking} & $73.7$ & $47.1$ & $41.0$ & $53.9$ & $54.0$ & $67.1$ & $58.5$ & $62.5$ \\
        ET$_{\mathtt{BERT}}$ \cite{gupta-durrett-2019-effective} & $73.6$ & $52.6$ & -- & -- & -- & -- & -- & -- \\
        \textsc{Dynapro} \cite{amini2020procedural} & $72.4$ & $49.3$ & $\mathbf{44.5}$ & $55.4$ & $55.5$ & $75.2$ & $58.0$ & $65.5$ \\
        TSLM \cite{rajaby-faghihi-kordjamshidi-2021-time} & $78.8$ & $56.8$ & $40.9$ & $58.8$ & $58.4$ & $68.4$ & $68.9$ & $68.6$ \\
        \textsc{KoaLa} \cite{zhang2021knowledge} & $78.5$ & $53.3$ & $41.3$ & $57.7$ & $57.5$ & $77.7$ & $64.4$ & $70.4$ \\
        REAL \cite{huang2021reasoning} & $78.4$ & $53.7$ & $42.4$ & $58.2$ & $57.9$ & $\mathbf{81.9}$ & $61.9$ & $70.5$ \\
        \midrule
        \ours & $\mathbf{81.7}$ & $\mathbf{58.3}$ & $43.3$ & $\mathbf{61.1}$ & $\mathbf{60.7}$ & $74.8$ & $\mathbf{69.8}$ & $\mathbf{72.2}$ \\
        ~~~~\textit{w.o.} pre-training & $78.8$ & $57.2$ & $42.9$ & $59.6$ & $59.2$ & $69.9$ & $68.1$ & $69.0$ \\
        \bottomrule
    \end{tabular}
    }
    \caption{Experimental results of our method \ours and baselines on the test set of \propara.}
    \label{tab:propara_result}
\end{table*}

\subsection{Data and Evaluation}

\paragraph{\alchemy, \scene \& \tangrams}
These three tasks are introduced with different environments in the \scone corpus \cite{long-etal-2016-simpler}.
Each human-agent interaction has $5$ instructions.
Following \citet{long-etal-2016-simpler}, we evaluate \ours with denotation accuracy.
In addition, the evaluation metrics can be divided into the denotation accuracy of a single instruction (Inst), of the first three instructions (3utts), and of the complete interactions (5utts).

\paragraph{\propara \& \recipes}

These two tasks is introduced in two procedural text understanding datasets \cite{dalvi-etal-2018-tracking,bosselut2017simulating}, and are designed to track entity states through natural language paragraphs.
For \propara, the evaluation metrics are composed of two levels: the sentence-level and the document-level.
The \textbf{sentence-level} evaluates the model based on its prediction for the following three questions: Is entity Created, Moved or Destroyed in the process? When is entity Created, Moved or Destroyed? Where is entity Created, Moved or Destroyed?
The sentence-level metrics include the accuracy of the above questions (Cat-1, Cat-2, Cat-3), and their micro\,/\,macro-average.
The \textbf{document-level} evaluates the model based on its prediction on four document-level questions: What are the inputs? What are the outputs? What are the conversions? What are the moves?
The document-level metrics report the average precision, recall, and F1 score of the four questions.
For \recipes, following previous work \cite{zhang2021knowledge,huang2021reasoning}, we report the location changes of each entity, and take precision, recall and F1 scores as the evaluation metrics.
The statistics of 5 datasets can be found in Appendix \ref{appendix-statistics}.

\begin{table}[]
    \centering
    \small
    \scalebox{0.95}{
    \begin{tabular}{p{3.6cm}ccc}
        \toprule
        \textbf{Models} & \textbf{Precision} & \textbf{Recall} & \textbf{F1} \\
        \midrule
        NCET (re-implementation) & $56.5$ & $46.4$ & $50.9$ \\
        IEN (re-implementation) & $58.5$ & $47.0$ & $52.2$ \\
        \textsc{KoaLa} \cite{zhang2021knowledge} & $\mathbf{60.1}$ & $52.6$ & $56.1$ \\
        REAL \cite{huang2021reasoning} & $55.2$ & $52.9$ & $54.1$ \\
        \midrule
        \ours & $56.0$ & $\mathbf{67.1}$ & $\mathbf{61.1}$ \\
        ~~~~\textit{w.o.} pre-training & $53.9$ & $63.6$ & $58.4$ \\
        \bottomrule
    \end{tabular}
    }
    \caption{Experimental results of our method \ours and baselines on the test set of \recipes.}
    \label{tab:recipes_result}
\end{table}

\subsection{Experimental Setup}

We use BART-Large in fairseq \cite{ott-etal-2019-fairseq} to implement \ours.
During pre-training, we synthesize $1$ million pre-training examples for each experimental task.
The learning rate is set to $3\times10^{-5}$ in all experiments of pre-training and fine-tuning.
During pre-training, the maximum training step is set to $10,000$ for \alchemy, \scene, \tangrams and $2,000$ for \propara and \recipes, while the batch size is set to around $1,000$ for all tasks.
During fine-tuning, the maximum training step is set to $10,000$ for all tasks, while the batch size is set to $64$ for \alchemy, \scene, \tangrams and $32$ for \propara and \recipes, respectively.

\subsection{Experimental Results}

\paragraph{\alchemy \& \scene}
From Table \ref{tab:scone_result}, we can observe that \ours outperforms previous best-performing systems under weak supervision on both \alchemy and \scene, with significant improvements of $13.1$\% and $5.9$\% in the 5utts denotation accuracy, respectively.
Notably, \ours not only achieves new state-of-the-art performance among weakly supervised approaches, but also comes close to the performance of fully supervised approaches that leverage extra annotated programs.
Moreover, the results also show that the execution-guided pre-training brings significant improvements (e.g., $4.1$\% of \alchemy in the 5utts denotation accuracy), which demonstrates that our pre-training strategy provides considerable prior knowledge for \ours.

\paragraph{\tangrams}
Similarly, the results on \tangrams in Table \ref{tab:scone_result} show that our execution-guided pre-training strategy improves \ours by $3.3$\% in the 5utts denotation accuracy, further illustrating the effectiveness of our approach.
Nevertheless, \ours does not perform as well compared to previous state-of-the-art method \cite{suhr-artzi-2018-situated}.
We suppose this is because \citet{suhr-artzi-2018-situated} carefully model the historical instructions, while \ours directly concatenates them.
We leave the fine-grained context modeling of our approach for future work.

\begin{figure}[t]
    \centering
    \includegraphics[width=0.4\textwidth]{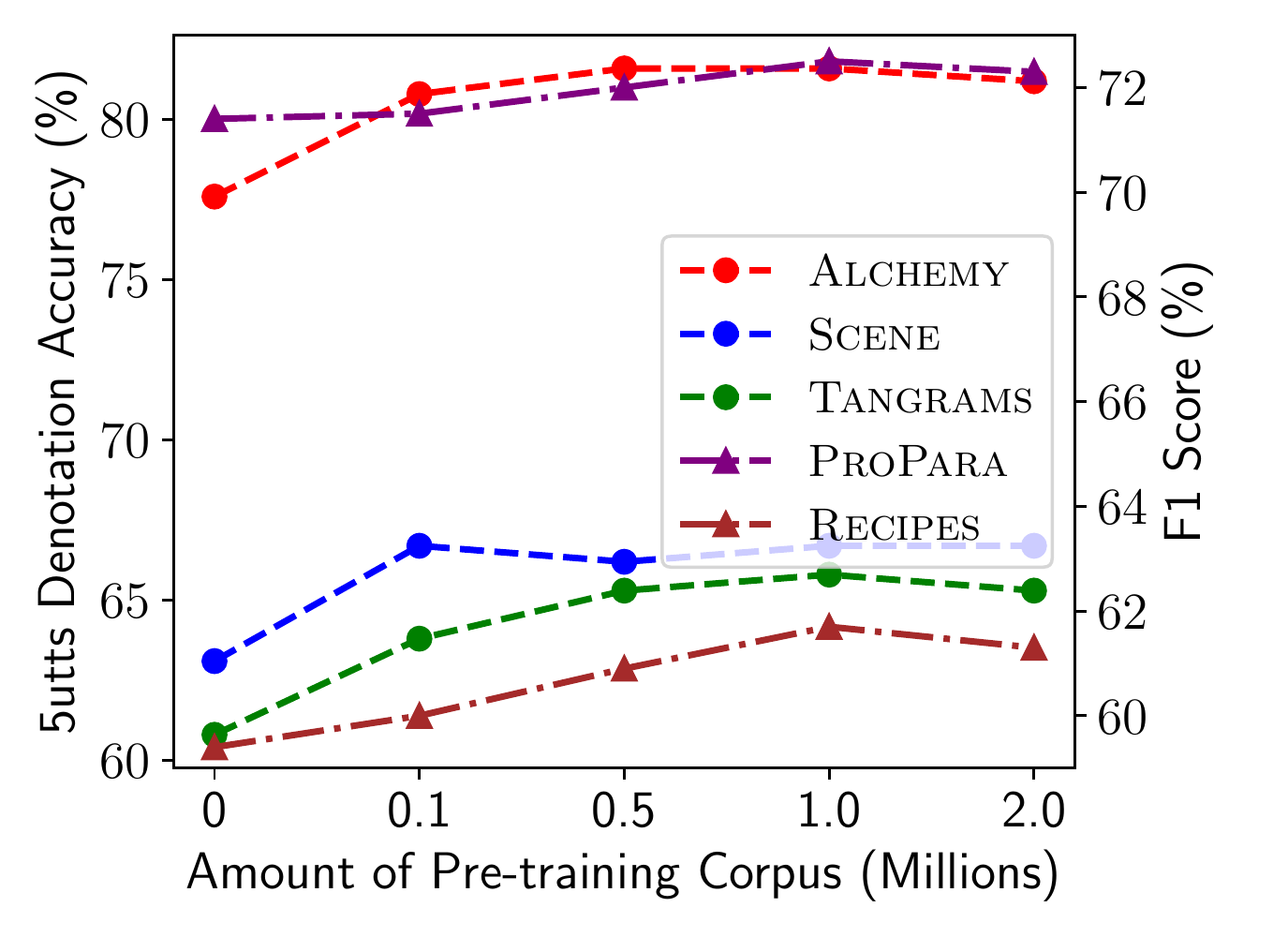}
    \caption{The performance of downstream tasks with respect to the amount of pre-training corpus. 
    We plot the 5utts denotation accuracy on \alchemy, \scene and \tangrams (circle), and  plot the F1 score on \propara and \recipes (triangle).}
    \label{fig:corpus_size}
\end{figure}

\paragraph{\propara}
Table \ref{tab:propara_result} summarizes the results of the \propara task, in which \ours achieves new state-of-the-art performance based on both the sentence-level and the document-level evaluation.
On the sentence-level evaluation, \ours shows stable improvements in most metrics compared to both previous approaches and \ours \textit{w.o.} execution-guided pre-training, which demonstrates that \ours achieves an overall improvement in the understanding of procedural texts with respect to the environment. 
On the document-level evaluation, \ours achieves an F1 score of $72.2$\%, which is $1.7$\% higher than the previous best-performing system REAL \cite{huang2021reasoning} and $1.8$\% higher than \textsc{KoaLa} \cite{zhang2021knowledge}.
Note the improvement is highly non-trivial since \textsc{KoaLa} leverages external knowledge, which indicates that the prior knowledge \ours learns during pre-training is more effective than external knowledge.
Similarly, the execution-guided pre-training brings a $3.2$\% improvement, which again demonstrates that the pre-training in \ours can significantly facilitate the interaction procedure between natural language and environments.

\paragraph{\recipes}
Table \ref{tab:recipes_result} shows the experimental results of the \recipes task that \ours reach state-of-the-art performance and surpass previous best-performing systems \cite{huang2021reasoning} with a large margin by $7.0$\%. Besides, the proposed execution-guided pre-training also brings a $2.7$\% improvement. These results further illustrate the effectiveness of \ours.

\subsection{Pre-training Analysis}

\paragraph{Scaling up pre-training has a positive impact}
Previous work \cite{lewis-etal-2020-bart} has shown that the scale of the pre-training corpus is an important factor in pre-training, and thus we analyze the effect of our pre-training scale on downstream tasks.
Figure \ref{fig:corpus_size} shows the performance of downstream tasks with respect to the size of the pre-training corpus, which are obtained from the validation set of each task.
As seen, the performance of the model generally improves by scaling up the pre-training corpus, consistent with previous observations on pre-training \cite{liu2021tapex}.


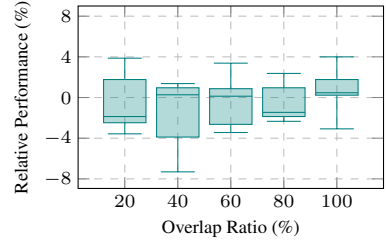
\begin{figure}[t]
    \centering
    \scriptsize
    \begin{tikzpicture}
\begin{axis}[
    boxplot/draw direction=y,
    width=.35\textwidth, 
    ymax=9,
    xlabel={{Overlap Ratio (\%)}},
    xlabel style={align=center,yshift=0em},
    ylabel={Relative Performance (\%)},
    height=.25\textwidth ,
    ytick={-8.0,-4.0,...,8.0},
    grid style=dashed,
    ymajorgrids=true,
    xmajorgrids=true,
    xtick={1,2,3,4,5},
    xticklabels={$20$, $40$, $60$, $80$, $100$},
    every axis plot/.append style={fill,fill opacity=0.3}
]
\addplot+[
boxplot prepared={
lower whisker=-3.57, lower quartile=-2.48,
median=-1.87,
upper quartile=1.77, upper whisker=3.87,
},
color=teal
]
coordinates {};
\addplot+[
boxplot prepared={
lower whisker=-7.31, lower quartile=-3.89,
median=0.27,
upper quartile=0.96, upper whisker=1.37,
},
color=teal
]
coordinates {};
\addplot+[
boxplot prepared={
lower whisker=-3.44, lower quartile=-2.64,
median=0.13,
upper quartile=0.87, upper whisker=3.38,
},
color=teal
]
coordinates {};
\addplot+[
boxplot prepared={
lower whisker=-2.34, lower quartile=-1.87,
median=-1.46,
upper quartile=0.96, upper whisker=2.37,
},
color=teal
]
coordinates {};
\addplot+[
boxplot prepared={
lower whisker=-3.08, lower quartile=0.23,
median=0.47,
upper quartile=1.77, upper whisker=4.00,
},
color=teal
]
coordinates {};
\end{axis}
\end{tikzpicture}
\caption{The relative performance of downstream tasks with respect to the overlap ratio. }
\label{fig:overlap_size}
\end{figure}


\begin{table*}[t]
    \centering
    \small
    \scalebox{0.85}{
    \begin{tabular}{p{2cm}p{5.8cm}p{9.5cm}}
        \toprule
        \textbf{Type (Percent)} & \textbf{Example} & \textbf{Environment (Goal State) Comparison} \\
        \midrule
        \multirow{2}{2cm}{Operation Correctness ($68.3$\%)} & \textbf{Environment (Initial State)}: \figinline{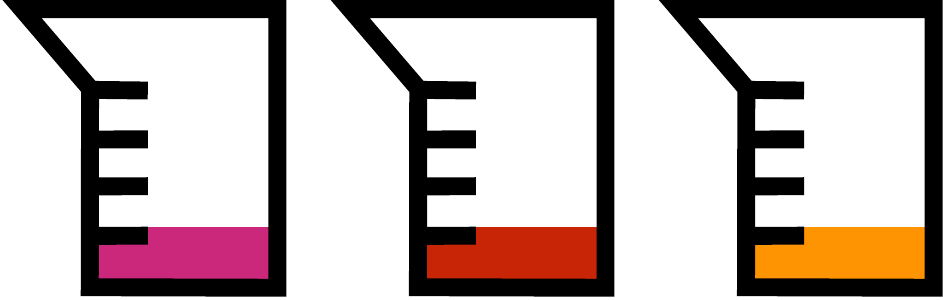} &  
        \multirow{2}{9.5cm}{The difference lies in the third beaker, \figinline{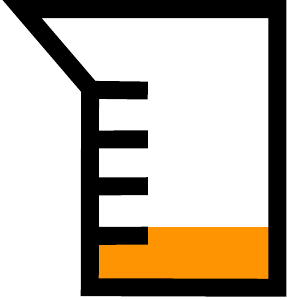} (\textit{w.o.} pre-training) versus \figinline{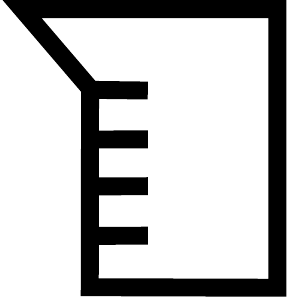} (\textit{w.} pre-training).
        Without pre-training, the model does not correctly understand the semantics of ``add'', i.e., removing the liquid from the third beaker.} 
        \\  & \textbf{Instruction}: Empty out the first beaker, add the orange chemical to the red. & 
        \\
        \midrule
        \multirow{2}{2cm}{Instruction Completeness ($18.3$\%)} & \textbf{Environment (Initial State)}: \figinline{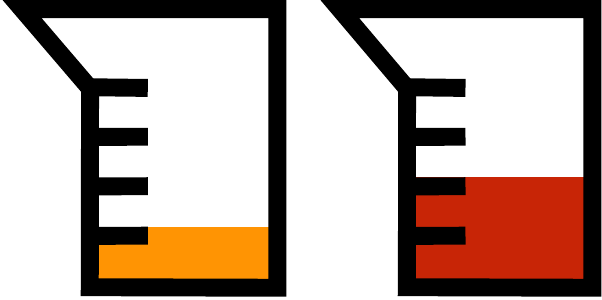} & 
         \multirow{2}{9.5cm}{The difference lies in the first beaker, \figinline{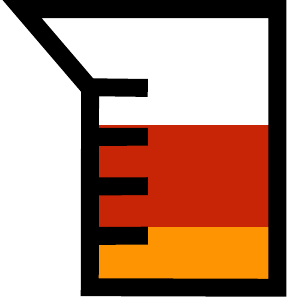} (\textit{w.o.} pre-training) versus \figinline{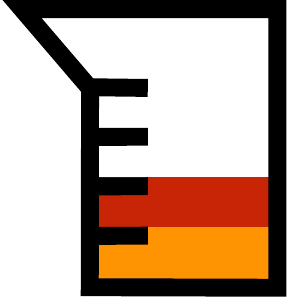} (\textit{w.} pre-training). As observed, without pre-training, the model seems to ignore the instruction ``throw out one unit of the second beaker''.} \\
         & \textbf{Instruction}: Throw out one unit of the second beaker, pour the second beaker into first one. & \\
        \midrule
        \multirow{2}{2.1cm}{Grounding Correctness ($8.5$\%)} & \textbf{Environment (Initial State)}: \figinline{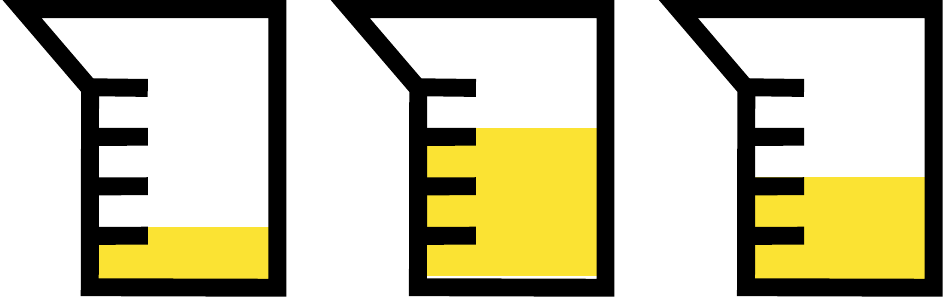} & 
        \multirow{2}{9.5cm}{The whole states are \figinline{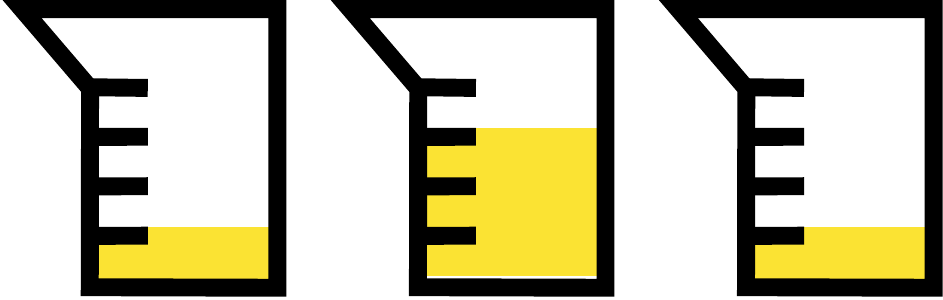} (\textit{w.o.} pre-training) versus \figinline{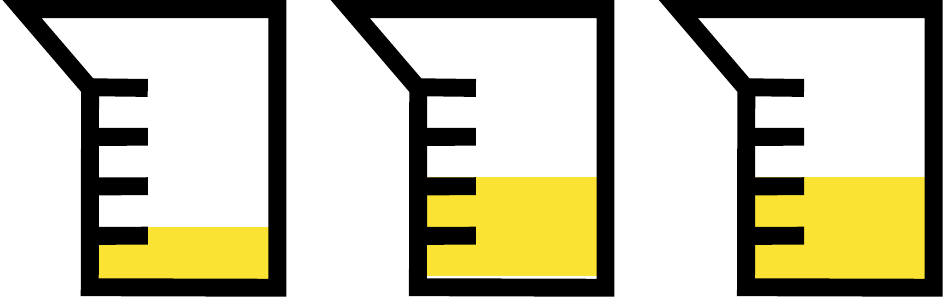} (\textit{w.} pre-training). Without pre-training, the model does not find the correct beaker according to the instructions.} \\
         & \textbf{Instruction}: Pour out one part of the second yellow beaker. & \\
        \bottomrule
    \end{tabular}
    }
    \caption{The main types of the improvements by the execution-guided pre-training in the validation set of \alchemy. For the all domains, please refer to Appendix \ref{sec:improvement-appendix}. }
    \label{tab:pretrain-improvement}
\end{table*}

\paragraph{Improvements do not come from data leakage}
Since the pre-training corpus of \ours contains various randomly sampled environment states, this may raise the doubt that the improvements of \ours is due to the data leakage, that is, \ours has seen some environments in the downstream validation sets. Although we have already ensure that the pre-training corpus does not contain the environment states in the validation set and the test set, it is still interesting to investigate the potential impact of data leakage on \ours.
To analyze the effect, we create a validation corpus of size $40,000$ for each task, which contains only the environment states in the validation set, and then merge the cases from the validation corpus into the pre-training corpus with certain ratios (denoted as \textbf{overlap ratio}).
Figure \ref{fig:overlap_size} shows the box plot of the relative performance to the reported performance (vertical axis) with respect to the overlap ratio (horizontal axis).
We can observe that from the perspective of the vertical axis, the vertical axis of the densest area is near 0, indicating that the two variables are irrelevant, thus proving that the effectiveness of \ours hardly relies on the overlap between corpus size and validation set.
For the detailed downstream performance with respect to the overlap ratio, please refer to Appendix \ref{dp-wrt-or}.

\paragraph{Improvements come from prior knowledge acquirement}
To show what \ours obtain from the execution-guided pre-training procedure, we manually analyze examples in the validation set where predictions are wrong before pre-training and correct after pre-training. Table \ref{tab:pretrain-improvement} shows the main types of the improvements caused by the execution-guided pre-training. 
We can see that with the execution-guided pre-training, \ours successfully masters the prior knowledge of different environments.
Specifically, \ours can manipulate the environments better, as reflected in the correctness of operations, the completeness of instructions, and the correctness of grounding.

\section{Related Work}


\paragraph{Language-based Environment Manipulation}
The first line of our related work is the previous work on LEM tasks.
According to the output, existing methods on LEM tasks can be mainly divided into two categories: program prediction and state prediction.
Prior work always treat the LEM task as a program prediction problem \cite{long-etal-2016-simpler,guu-etal-2017-language,suhr-artzi-2018-situated,fried-etal-2018-unified,huang2018flowqa,yeh2019flowdelta,dalvi-etal-2019-everything}. 
However, these approaches are environment-dependent and cannot be easily adapted to other environments. Besides, they either rely on natural language-program pairs as supervision or require complex heuristic rules, which is costly.
Recent approaches generally treat the LEM task as a state prediction problem by predicting the goal state directly \cite{dalvi-etal-2018-tracking,du-etal-2019-consistent,das2018building,tang-etal-2020-understanding-procedural,rajaby-faghihi-kordjamshidi-2021-time,zhang2021knowledge}. 
These models can eliminate the data collection issue, but require complex models designed to meet the needs of different kinds of environments.
Compared with the above work, \ours has the following advantages: 
1) The proposed task-agnostic approach does not require additional annotations and is easy to generalize across different environments. 2) The proposed execution-guided pre-training strategy can further improve the model performance with synthetic data only.

\paragraph{Program Execution}
The second line of our related work is the execution-guided work, of which the most related work are ProTo \cite{zhao2021proto} and \textsc{TaPEx} \cite{liu2021tapex}.
ProTo learns to execute given programs on the observed task specifications, which focuses on following a given program to perform the corresponding task.
Different from ProTo, \ours focuses on pre-training with program execution to enhance the downstream performance.
Following a similar idea, \textsc{TaPEx} \cite{liu2021tapex} improves the table pre-training by learning SQL execution over tables.
The main difference between \textsc{TaPEx} and \ours is that \textsc{TaPEx} choose SQL execution as the pre-training task, which is suitable for a single environment only. While, \ours is more flexible, and enables us to systematically design the pre-training task and synthesize pre-training corpus based on environment properties, and proven effective on multiple environments.

\section{Conclusion \& Future Work}
In this work, we propose \ours, a general framework for language-based environment manipulation tasks that not only models different environments using the same protocol, but also injects prior knowledge of environments into our model. 
Experimental results on five tasks demonstrate the effectiveness of \ours: the execution-guided pre-training strategy brings significant improvements on all of them and \ours achieves the state-of-the-art performance on four of them.
For future work, we hope to extend our approach to more complex environments and tasks such as image editing \cite{fu-etal-2020-sscr} and text editing \cite{faltings-etal-2021-text}.

\section*{Limitations}
The main limitation in this paper is that \ours focus on symbolic environments instead of raw environments with only visual features. Compared to the latter, the former can be represented by semantic symbols, and thus enjoys better controllability and interpretability. We leave the exploration of raw environments for future work.

\section*{Ethics Statement}
In this paper, we propose \ours, a general framework for language-based environment manipulation tasks, consisting of a task-agnostic approach and an execution-guided pre-training strategy. We conduct experiments on five benchmarks, namely, \alchemy, \scene, \tangrams, \propara, \recipes. All benchmarks are free and open for research use. The pre-training corpus is generated based on open-source program grammars, which are no ethics issues. 

\section*{Acknowledgement}
We would like to thank the anonymous reviewers for their helpful comments. 

\bibliography{anthology,custom}

\begin{thebibliography}{35}
\expandafter\ifx\csname natexlab\endcsname\relax\def\natexlab#1{#1}\fi

\bibitem[{Amini et~al.(2020)Amini, Bosselut, Mishra, Choi, and
  Hajishirzi}]{amini2020procedural}
Aida Amini, Antoine Bosselut, Bhavana~Dalvi Mishra, Yejin Choi, and Hannaneh
  Hajishirzi. 2020.
\newblock \href {https://doi.org/10.24432/C5C883} {Procedural reading
  comprehension with attribute-aware context flow}.
\newblock In \emph{Conference on Automated Knowledge Base Construction, {AKBC}
  2020, Virtual, June 22-24, 2020}.

\bibitem[{Andreas and Klein(2015)}]{andreas-klein-2015-alignment}
Jacob Andreas and Dan Klein. 2015.
\newblock \href {https://doi.org/10.18653/v1/D15-1138} {Alignment-based
  compositional semantics for instruction following}.
\newblock In \emph{Proceedings of the 2015 Conference on Empirical Methods in
  Natural Language Processing}, pages 1165--1174, Lisbon, Portugal. Association
  for Computational Linguistics.

\bibitem[{Bosselut et~al.(2018)Bosselut, Levy, Holtzman, Ennis, Fox, and
  Choi}]{bosselut2017simulating}
Antoine Bosselut, Omer Levy, Ari Holtzman, Corin Ennis, Dieter Fox, and Yejin
  Choi. 2018.
\newblock \href {https://openreview.net/forum?id=rJYFzMZC-} {Simulating action
  dynamics with neural process networks}.
\newblock In \emph{6th International Conference on Learning Representations,
  {ICLR} 2018, Vancouver, BC, Canada, April 30 - May 3, 2018, Conference Track
  Proceedings}. OpenReview.net.

\bibitem[{Dalvi et~al.(2018)Dalvi, Huang, Tandon, Yih, and
  Clark}]{dalvi-etal-2018-tracking}
Bhavana Dalvi, Lifu Huang, Niket Tandon, Wen-tau Yih, and Peter Clark. 2018.
\newblock \href {https://doi.org/10.18653/v1/N18-1144} {Tracking state changes
  in procedural text: a challenge dataset and models for process paragraph
  comprehension}.
\newblock In \emph{Proceedings of the 2018 Conference of the North {A}merican
  Chapter of the Association for Computational Linguistics: Human Language
  Technologies, Volume 1 (Long Papers)}, pages 1595--1604, New Orleans,
  Louisiana. Association for Computational Linguistics.

\bibitem[{Dalvi et~al.(2019)Dalvi, Tandon, Bosselut, Yih, and
  Clark}]{dalvi-etal-2019-everything}
Bhavana Dalvi, Niket Tandon, Antoine Bosselut, Wen-tau Yih, and Peter Clark.
  2019.
\newblock \href {https://doi.org/10.18653/v1/D19-1457} {Everything happens for
  a reason: Discovering the purpose of actions in procedural text}.
\newblock In \emph{Proceedings of the 2019 Conference on Empirical Methods in
  Natural Language Processing and the 9th International Joint Conference on
  Natural Language Processing (EMNLP-IJCNLP)}, pages 4496--4505, Hong Kong,
  China. Association for Computational Linguistics.

\bibitem[{Das et~al.(2019)Das, Munkhdalai, Yuan, Trischler, and
  McCallum}]{das2018building}
Rajarshi Das, Tsendsuren Munkhdalai, Xingdi Yuan, Adam Trischler, and Andrew
  McCallum. 2019.
\newblock \href {https://openreview.net/forum?id=S1lhbnRqF7} {Building dynamic
  knowledge graphs from text using machine reading comprehension}.
\newblock In \emph{7th International Conference on Learning Representations,
  {ICLR} 2019, New Orleans, LA, USA, May 6-9, 2019}. OpenReview.net.

\bibitem[{Du et~al.(2019)Du, Dalvi, Tandon, Bosselut, Yih, Clark, and
  Cardie}]{du-etal-2019-consistent}
Xinya Du, Bhavana Dalvi, Niket Tandon, Antoine Bosselut, Wen-tau Yih, Peter
  Clark, and Claire Cardie. 2019.
\newblock \href {https://doi.org/10.18653/v1/N19-1244} {Be consistent!
  improving procedural text comprehension using label consistency}.
\newblock In \emph{Proceedings of the 2019 Conference of the North {A}merican
  Chapter of the Association for Computational Linguistics: Human Language
  Technologies, Volume 1 (Long and Short Papers)}, pages 2347--2356,
  Minneapolis, Minnesota. Association for Computational Linguistics.

\bibitem[{Faltings et~al.(2021)Faltings, Galley, Hintz, Brockett, Quirk, Gao,
  and Dolan}]{faltings-etal-2021-text}
Felix Faltings, Michel Galley, Gerold Hintz, Chris Brockett, Chris Quirk,
  Jianfeng Gao, and Bill Dolan. 2021.
\newblock \href {https://doi.org/10.18653/v1/2021.naacl-main.414} {Text editing
  by command}.
\newblock In \emph{Proceedings of the 2021 Conference of the North American
  Chapter of the Association for Computational Linguistics: Human Language
  Technologies}, pages 5259--5274, Online. Association for Computational
  Linguistics.

\bibitem[{Fried et~al.(2018)Fried, Andreas, and
  Klein}]{fried-etal-2018-unified}
Daniel Fried, Jacob Andreas, and Dan Klein. 2018.
\newblock \href {https://doi.org/10.18653/v1/N18-1177} {Unified pragmatic
  models for generating and following instructions}.
\newblock In \emph{Proceedings of the 2018 Conference of the North {A}merican
  Chapter of the Association for Computational Linguistics: Human Language
  Technologies, Volume 1 (Long Papers)}, pages 1951--1963, New Orleans,
  Louisiana. Association for Computational Linguistics.

\bibitem[{Fu et~al.(2020)Fu, Wang, Grafton, Eckstein, and
  Wang}]{fu-etal-2020-sscr}
Tsu-Jui Fu, Xin Wang, Scott Grafton, Miguel Eckstein, and William~Yang Wang.
  2020.
\newblock \href {https://doi.org/10.18653/v1/2020.emnlp-main.357} {{SSCR}:
  Iterative language-based image editing via self-supervised counterfactual
  reasoning}.
\newblock In \emph{Proceedings of the 2020 Conference on Empirical Methods in
  Natural Language Processing (EMNLP)}, pages 4413--4422, Online. Association
  for Computational Linguistics.

\bibitem[{Gupta and Durrett(2019{\natexlab{a}})}]{gupta-durrett-2019-effective}
Aditya Gupta and Greg Durrett. 2019{\natexlab{a}}.
\newblock \href {https://doi.org/10.18653/v1/D19-1070} {Effective use of
  transformer networks for entity tracking}.
\newblock In \emph{Proceedings of the 2019 Conference on Empirical Methods in
  Natural Language Processing and the 9th International Joint Conference on
  Natural Language Processing (EMNLP-IJCNLP)}, pages 759--769, Hong Kong,
  China. Association for Computational Linguistics.

\bibitem[{Gupta and Durrett(2019{\natexlab{b}})}]{gupta-durrett-2019-tracking}
Aditya Gupta and Greg Durrett. 2019{\natexlab{b}}.
\newblock \href {https://doi.org/10.18653/v1/W19-1502} {Tracking discrete and
  continuous entity state for process understanding}.
\newblock In \emph{Proceedings of the Third Workshop on Structured Prediction
  for {NLP}}, pages 7--12, Minneapolis, Minnesota. Association for
  Computational Linguistics.

\bibitem[{Guu et~al.(2017)Guu, Pasupat, Liu, and
  Liang}]{guu-etal-2017-language}
Kelvin Guu, Panupong Pasupat, Evan Liu, and Percy Liang. 2017.
\newblock \href {https://doi.org/10.18653/v1/P17-1097} {From language to
  programs: Bridging reinforcement learning and maximum marginal likelihood}.
\newblock In \emph{Proceedings of the 55th Annual Meeting of the Association
  for Computational Linguistics (Volume 1: Long Papers)}, pages 1051--1062,
  Vancouver, Canada. Association for Computational Linguistics.

\bibitem[{Henaff et~al.(2017)Henaff, Weston, Szlam, Bordes, and
  LeCun}]{henaff2016tracking}
Mikael Henaff, Jason Weston, Arthur Szlam, Antoine Bordes, and Yann LeCun.
  2017.
\newblock \href {https://openreview.net/forum?id=rJTKKKqeg} {Tracking the world
  state with recurrent entity networks}.
\newblock In \emph{5th International Conference on Learning Representations,
  {ICLR} 2017, Toulon, France, April 24-26, 2017, Conference Track
  Proceedings}. OpenReview.net.

\bibitem[{Huang et~al.(2021)Huang, Geng, Pei, Long, and
  Jiang}]{huang2021reasoning}
Hao Huang, Xiubo Geng, Jian Pei, Guodong Long, and Daxin Jiang. 2021.
\newblock Reasoning over entity-action-location graph for procedural text
  understanding.
\newblock In \emph{Proceedings of the 59th Annual Meeting of the Association
  for Computational Linguistics and the 11th International Joint Conference on
  Natural Language Processing (Volume 1: Long Papers)}, pages 5100--5109.

\bibitem[{Huang et~al.(2019)Huang, Choi, and Yih}]{huang2018flowqa}
Hsin{-}Yuan Huang, Eunsol Choi, and Wen{-}tau Yih. 2019.
\newblock \href {https://openreview.net/forum?id=ByftGnR9KX} {Flowqa: Grasping
  flow in history for conversational machine comprehension}.
\newblock In \emph{7th International Conference on Learning Representations,
  {ICLR} 2019, New Orleans, LA, USA, May 6-9, 2019}. OpenReview.net.

\bibitem[{Lewis et~al.(2020)Lewis, Liu, Goyal, Ghazvininejad, Mohamed, Levy,
  Stoyanov, and Zettlemoyer}]{lewis-etal-2020-bart}
Mike Lewis, Yinhan Liu, Naman Goyal, Marjan Ghazvininejad, Abdelrahman Mohamed,
  Omer Levy, Veselin Stoyanov, and Luke Zettlemoyer. 2020.
\newblock \href {https://doi.org/10.18653/v1/2020.acl-main.703} {{BART}:
  Denoising sequence-to-sequence pre-training for natural language generation,
  translation, and comprehension}.
\newblock In \emph{Proceedings of the 58th Annual Meeting of the Association
  for Computational Linguistics}, pages 7871--7880, Online. Association for
  Computational Linguistics.

\bibitem[{Liu et~al.(2020)Liu, Chen, Guo, Lou, Zhou, and Zhang}]{qian2020how}
Qian Liu, Bei Chen, Jiaqi Guo, Jian{-}Guang Lou, Bin Zhou, and Dongmei Zhang.
  2020.
\newblock \href {https://doi.org/10.24963/ijcai.2020/495} {How far are we from
  effective context modeling? an exploratory study on semantic parsing in
  context}.
\newblock In \emph{Proceedings of the Twenty-Ninth International Joint
  Conference on Artificial Intelligence, {IJCAI} 2020}, pages 3580--3586.
  ijcai.org.

\bibitem[{Liu et~al.(2021)Liu, Chen, Guo, Ziyadi, Lin, Chen, and
  Lou}]{liu2021tapex}
Qian Liu, Bei Chen, Jiaqi Guo, Morteza Ziyadi, Zeqi Lin, Weizhu Chen, and
  Jian{-}Guang Lou. 2021.
\newblock \href {http://arxiv.org/abs/2107.07653} {{TAPEX:} table pre-training
  via learning a neural {SQL} executor}.
\newblock \emph{CoRR}, abs/2107.07653.

\bibitem[{Long et~al.(2016)Long, Pasupat, and Liang}]{long-etal-2016-simpler}
Reginald Long, Panupong Pasupat, and Percy Liang. 2016.
\newblock \href {https://doi.org/10.18653/v1/P16-1138} {Simpler
  context-dependent logical forms via model projections}.
\newblock In \emph{Proceedings of the 54th Annual Meeting of the Association
  for Computational Linguistics (Volume 1: Long Papers)}, pages 1456--1465,
  Berlin, Germany. Association for Computational Linguistics.

\bibitem[{Narayan-Chen et~al.(2019)Narayan-Chen, Jayannavar, and
  Hockenmaier}]{narayan-chen-etal-2019-collaborative}
Anjali Narayan-Chen, Prashant Jayannavar, and Julia Hockenmaier. 2019.
\newblock \href {https://doi.org/10.18653/v1/P19-1537} {Collaborative dialogue
  in {M}inecraft}.
\newblock In \emph{Proceedings of the 57th Annual Meeting of the Association
  for Computational Linguistics}, pages 5405--5415, Florence, Italy.
  Association for Computational Linguistics.

\bibitem[{Ott et~al.(2019)Ott, Edunov, Baevski, Fan, Gross, Ng, Grangier, and
  Auli}]{ott-etal-2019-fairseq}
Myle Ott, Sergey Edunov, Alexei Baevski, Angela Fan, Sam Gross, Nathan Ng,
  David Grangier, and Michael Auli. 2019.
\newblock \href {https://doi.org/10.18653/v1/N19-4009} {fairseq: A fast,
  extensible toolkit for sequence modeling}.
\newblock In \emph{Proceedings of the 2019 Conference of the North {A}merican
  Chapter of the Association for Computational Linguistics (Demonstrations)},
  pages 48--53, Minneapolis, Minnesota. Association for Computational
  Linguistics.

\bibitem[{Rajaby~Faghihi and
  Kordjamshidi(2021)}]{rajaby-faghihi-kordjamshidi-2021-time}
Hossein Rajaby~Faghihi and Parisa Kordjamshidi. 2021.
\newblock \href {https://doi.org/10.18653/v1/2021.naacl-main.362} {Time-stamped
  language model: Teaching language models to understand the flow of events}.
\newblock In \emph{Proceedings of the 2021 Conference of the North American
  Chapter of the Association for Computational Linguistics: Human Language
  Technologies}, pages 4560--4570, Online. Association for Computational
  Linguistics.

\bibitem[{Ribeiro et~al.(2019)Ribeiro, Hinrichs, Crouse, Forbus, Chang, and
  Witbrock}]{ribeiro2019predicting}
Danilo Ribeiro, Thomas Hinrichs, Maxwell Crouse, Kenneth Forbus, Maria Chang,
  and Michael Witbrock. 2019.
\newblock Predicting state changes in procedural text using analogical question
  answering.
\newblock In \emph{Proc. of ACS}.

\bibitem[{Seo et~al.(2017)Seo, Min, Farhadi, and Hajishirzi}]{seo2016query}
Min~Joon Seo, Sewon Min, Ali Farhadi, and Hannaneh Hajishirzi. 2017.
\newblock \href {https://openreview.net/forum?id=B1MRcPclx} {Query-reduction
  networks for question answering}.
\newblock In \emph{5th International Conference on Learning Representations,
  {ICLR} 2017, Toulon, France, April 24-26, 2017, Conference Track
  Proceedings}. OpenReview.net.

\bibitem[{Suhr and Artzi(2018)}]{suhr-artzi-2018-situated}
Alane Suhr and Yoav Artzi. 2018.
\newblock \href {https://doi.org/10.18653/v1/P18-1193} {Situated mapping of
  sequential instructions to actions with single-step reward observation}.
\newblock In \emph{Proceedings of the 56th Annual Meeting of the Association
  for Computational Linguistics (Volume 1: Long Papers)}, pages 2072--2082,
  Melbourne, Australia. Association for Computational Linguistics.

\bibitem[{Suhr et~al.(2019)Suhr, Yan, Schluger, Yu, Khader, Mouallem, Zhang,
  and Artzi}]{suhr-etal-2019-executing}
Alane Suhr, Claudia Yan, Jack Schluger, Stanley Yu, Hadi Khader, Marwa
  Mouallem, Iris Zhang, and Yoav Artzi. 2019.
\newblock \href {https://doi.org/10.18653/v1/D19-1218} {Executing instructions
  in situated collaborative interactions}.
\newblock In \emph{Proceedings of the 2019 Conference on Empirical Methods in
  Natural Language Processing and the 9th International Joint Conference on
  Natural Language Processing (EMNLP-IJCNLP)}, pages 2119--2130, Hong Kong,
  China. Association for Computational Linguistics.

\bibitem[{Tandon et~al.(2018)Tandon, Dalvi, Grus, Yih, Bosselut, and
  Clark}]{tandon-etal-2018-reasoning}
Niket Tandon, Bhavana Dalvi, Joel Grus, Wen-tau Yih, Antoine Bosselut, and
  Peter Clark. 2018.
\newblock \href {https://doi.org/10.18653/v1/D18-1006} {Reasoning about actions
  and state changes by injecting commonsense knowledge}.
\newblock In \emph{Proceedings of the 2018 Conference on Empirical Methods in
  Natural Language Processing}, pages 57--66, Brussels, Belgium. Association
  for Computational Linguistics.

\bibitem[{Tandon et~al.(2020)Tandon, Sakaguchi, Dalvi, Rajagopal, Clark,
  Guerquin, Richardson, and Hovy}]{tandon-etal-2020-dataset}
Niket Tandon, Keisuke Sakaguchi, Bhavana Dalvi, Dheeraj Rajagopal, Peter Clark,
  Michal Guerquin, Kyle Richardson, and Eduard Hovy. 2020.
\newblock \href {https://doi.org/10.18653/v1/2020.emnlp-main.520} {A dataset
  for tracking entities in open domain procedural text}.
\newblock In \emph{Proceedings of the 2020 Conference on Empirical Methods in
  Natural Language Processing (EMNLP)}, pages 6408--6417, Online. Association
  for Computational Linguistics.

\bibitem[{Tang et~al.(2020)Tang, Feng, and
  Zhao}]{tang-etal-2020-understanding-procedural}
Jizhi Tang, Yansong Feng, and Dongyan Zhao. 2020.
\newblock \href {https://doi.org/10.18653/v1/2020.emnlp-main.591}
  {Understanding procedural text using interactive entity networks}.
\newblock In \emph{Proceedings of the 2020 Conference on Empirical Methods in
  Natural Language Processing (EMNLP)}, pages 7281--7290, Online. Association
  for Computational Linguistics.

\bibitem[{Winograd(1971)}]{winograd1971procedures}
Terry Winograd. 1971.
\newblock Procedures as a representation for data in a computer program for
  understanding natural language.
\newblock Technical report.

\bibitem[{Yeh and Chen(2019)}]{yeh2019flowdelta}
Yi{-}Ting Yeh and Yun{-}Nung Chen. 2019.
\newblock \href {https://doi.org/10.18653/v1/D19-5812} {Flowdelta: Modeling
  flow information gain in reasoning for conversational machine comprehension}.
\newblock In \emph{Proceedings of the 2nd Workshop on Machine Reading for
  Question Answering, MRQA@EMNLP 2019, Hong Kong, China, November 4, 2019},
  pages 86--90. Association for Computational Linguistics.

\bibitem[{Zhang et~al.(2021)Zhang, Geng, Qin, Wu, and
  Jiang}]{zhang2021knowledge}
Zhihan Zhang, Xiubo Geng, Tao Qin, Yunfang Wu, and Daxin Jiang. 2021.
\newblock \href {https://doi.org/10.1145/3442381.3450126} {Knowledge-aware
  procedural text understanding with multi-stage training}.
\newblock In \emph{{WWW} '21: The Web Conference 2021, Virtual Event /
  Ljubljana, Slovenia, April 19-23, 2021}, pages 3512--3523. {ACM} / {IW3C2}.

\bibitem[{Zhao et~al.(2021)Zhao, Samel, Chen, and Song}]{zhao2021proto}
Zelin Zhao, Karan Samel, Binghong Chen, and Le~Song. 2021.
\newblock \href {http://arxiv.org/abs/2110.00804} {Proto: Program-guided
  transformer for program-guided tasks}.
\newblock \emph{CoRR}, abs/2110.00804.

\bibitem[{Zhong et~al.(2020)Zhong, Tang, Duan, Zhou, Wang, and
  Yin}]{zhong2020heterogeneous}
Wanjun Zhong, Duyu Tang, Nan Duan, Ming Zhou, Jiahai Wang, and Jian Yin. 2020.
\newblock \href {http://arxiv.org/abs/2004.12057} {A heterogeneous graph with
  factual, temporal and logical knowledge for question answering over dynamic
  contexts}.
\newblock \emph{CoRR}, abs/2004.12057.

\end{thebibliography}
\bibliographystyle{acl_natbib}

\appendix

\section{Program Grammar in Each Domain}
\label{all-grammar}
Table \ref{tab:alchemy_grammer}, Table \ref{tab:scene_grammar}, Table \ref{tab:tangrams_grammar} and Table \ref{tab:propara_recipes_grammar} show the grammar rules of used programs in each domain. 

\begin{table*}[t]
    \small
    \centering
    \begin{tabular}{rlp{8.5cm}}
         \toprule
          \multicolumn{2}{l}{\textbf{Grammar Rule}}  & \multicolumn{1}{l}{\textbf{Description}} \\
         \midrule
         $\langle$state$\rangle$ $~~\to$ & $\langle$action$\rangle$ $\langle$state$\rangle$ | $\langle$action$\rangle$ & A list of actions. \\
        $\langle$action$\rangle$  $~~\to$ & $\langle$mix$\rangle$ | $\langle$pour$\rangle$ | $\langle$drain$\rangle$ & One of the three actions. \\
          $\langle$mix$\rangle$ $~~\to$ & \textsc{Mix} ($\langle$beaker$\rangle$) & Mix the liquid in the $\langle$beaker$\rangle$ beaker. \\
          $\langle$pour$\rangle$ $~~\to$ & \textsc{Pour} ($\langle$beaker$\rangle$, $\langle$beaker$\rangle$) & Pour the liquid from the first beaker to the second beaker. \\
         $\langle$drain$\rangle$ $~~\to$ & \textsc{Drain} ($\langle$beaker$\rangle$, $\langle$integer$\rangle$) ~| & Pour out the $\langle$integer$\rangle$ unit from the $\langle$beaker$\rangle$ beaker. \\
         & \textsc{Drain} ($\langle$beaker$\rangle$, $\langle$fraction$\rangle$) & Pour $\langle$fraction$\rangle$ of the liquid out of the $\langle$beaker$\rangle$ beaker. \\ 
          $\langle$beaker$\rangle$ $~~\to$ & \textsc{Beaker} ($\langle$index$\rangle$) ~| & The $\langle$index$\rangle$-th beaker. \\
          & \textsc{Beaker} ($\langle$index$\rangle$,  $\langle$color$\rangle$) & The $\langle$index$\rangle$-th beaker of $\langle$color$\rangle$ color. \\
          $\langle$index$\rangle$ $~~\to$ & $1$ | $2$ |  $\cdots$ | $7$ | $-1$ | $-2$ | $\cdots$ | $-7$ & The index of the certain components in the environment. \\
         $\langle$color$\rangle$ $~~\to$ & r | g | o | p | y | b & The symbols corresponding to the color red, green, orange, purple, yellow and brown. \\
         $\langle$integer$\rangle$ $~~\to$ & $1$ | $2$ | $3$ | $4$ & The unit of the liquid. \\
         $\langle$fraction$\rangle$ $~~\to$ & $\frac{1}{2}$ | $\frac{1}{3}$ | $\frac{1}{4}$ | $\frac{2}{3}$ | $\frac{2}{4}$ | $\frac{3}{4}$ & The percentage of the liquid. \\
         \bottomrule
    \end{tabular}
    \caption{Grammar rules and corresponding descriptions of used program in \alchemy.}
    \label{tab:alchemy_grammer}
\end{table*}

\begin{table*}[t]
    \small
    \centering
    \begin{tabular}{rlp{7.5cm}}
         \toprule
         \multicolumn{2}{l}{\textbf{Grammar Rule}}  & \multicolumn{1}{l}{\textbf{Description}} \\
         \midrule
         $\langle$state$\rangle$ $~~\to$ & $\langle$action$\rangle$ $\langle$state$\rangle$ | $\langle$action$\rangle$ & A list of actions. \\
         $\langle$action$\rangle$  $~~\to$ & $\langle$person$\rangle$ | $\langle$rmperson$\rangle$ | $\langle$hat$\rangle$ | $\langle$rmhat$\rangle$ & One of the four actions. \\
         $\langle$person$\rangle$ $~~\to$ & \textsc{Person} ($\langle$index$\rangle$, $\langle$color$\rangle$) & Add a person with $\langle$color$\rangle$ shirt on the $\langle$index$\rangle$-th position. \\
         $\langle$rmperson$\rangle$ $~~\to$ & \textsc{RmPerson} ($\langle$index$\rangle$) & Remove the person on the $\langle$index$\rangle$-th position. \\
         $\langle$hat$\rangle$ $~~\to$ & \textsc{Hat} ($\langle$index$\rangle$, $\langle$color$\rangle$) & Add a hat of $\langle$color$\rangle$ color for the person on the $\langle$index$\rangle$-th position. \\
         $\langle$rmhat$\rangle$ $~~\to$ & \textsc{RmHat} ($\langle$index$\rangle$) & Remove the person's hat on the $\langle$index$\rangle$-th position. \\
         $\langle$index$\rangle$ $~~\to$ & $1$ | $2$ | $3$ | $\cdots$ | $10$ & The index of the certain components in the environment. \\
         $\langle$color$\rangle$ $~~\to$ & r | g | o | p | y | b & The symbols corresponding to the color red, green, orange, purple, yellow and brown. \\
         \bottomrule

    \end{tabular}
    \caption{Grammar rules and corresponding descriptions of used program in \scene.}
    \label{tab:scene_grammar}
\end{table*}

\begin{table*}[t]
    \small
    \centering
    \begin{tabular}{rlp{8.5cm}}
         \toprule
         \multicolumn{2}{l}{\textbf{Grammar Rule}}  & \multicolumn{1}{l}{\textbf{Description}} \\
         \midrule
         $\langle$state$\rangle$ $~~\to$ & $\langle$action$\rangle$ $\langle$state$\rangle$ | $\langle$action$\rangle$ & A list of actions. \\
         $\langle$action$\rangle$  $~~\to$ & $\langle$insert$\rangle$ | $\langle$remove$\rangle$ & One of the two actions. \\
         $\langle$insert$\rangle$ $~~\to$ & \textsc{Insert} ($\langle$index$\rangle$, $\langle$object$\rangle$) & Insert the $\langle$object$\rangle$ object at the $\langle$index$\rangle$ position. \\
         $\langle$remove$\rangle$ $~~\to$ & \textsc{Remove} ($\langle$index$\rangle$) & Remove the object at the $\langle$index$\rangle$ position. \\
         $\langle$index$\rangle$ $~~\to$ & $1$ | $2$ | $3$ | $4$ | $5$ & The index of the certain components in the environment. \\
         $\langle$object$\rangle$ $~~\to$ & A | B | C | D | E & The object name. \\
        \bottomrule

    \end{tabular}
    \caption{Grammar rules and corresponding descriptions of used program in \tangrams.}
    \label{tab:tangrams_grammar}
\end{table*}

\begin{table*}[t]
    \small
    \centering
    \begin{tabular}{rlp{7cm}}
         \toprule
         \multicolumn{2}{l}{\textbf{Grammar Rule}}  & \multicolumn{1}{l}{\textbf{Description}} \\
         \midrule
         $\langle$state$\rangle$ $~~\to$ & $\langle$action$\rangle$ $\langle$state$\rangle$ | $\langle$action$\rangle$ & A list of actions. \\
         $\langle$action$\rangle$  $~~\to$ & $\langle$create$\rangle$ | $\langle$move$\rangle$ | $\langle$destroy$\rangle$ & One of the three actions. \\
         $\langle$create$\rangle$ $~~\to$ & \textsc{Create} ($\langle$participant$\rangle$, $\langle$location$\rangle$) & Create $\langle$participant$\rangle$ at the $\langle$location$\rangle$. \\
         & \textsc{Create} ($\langle$participant$\rangle$, $?$) & Do not fill the location if $\langle$location$\rangle$ is not given. \\
         $\langle$move$\rangle$ $~~\to$ & \textsc{Move} ($\langle$participant$\rangle$, $\langle$location1$\rangle$, $\langle$location2$\rangle$) & Move $\langle$participant$\rangle$ from $\langle$location1$\rangle$ to $\langle$location2$\rangle$. \\
         $\langle$destroy$\rangle$ $~~\to$ & \textsc{Destroy} ($\langle$participant$\rangle$) & Remove $\langle$participant$\rangle$ from the current location. \\
         $\langle$participants$\rangle$ $~~\to$ & water | light | carbon | ... & Entities in the environments. \\
         $\langle$locations$\rangle$ $~~\to$ & soil | sun | cloud | ... & Entities' locations in the environments \\
         \bottomrule

    \end{tabular}
    \caption{Grammar rules and corresponding descriptions of used program in \propara and \recipes.}
    \label{tab:propara_recipes_grammar}
\end{table*}

\section{Statistics of Each Dataset}
\label{appendix-statistics}
Table \ref{tab:statistics} show the data statistics for \alchemy, \scene, \tangrams, \propara, \recipes. 

\begin{table*}[t]
    \small
    \centering
    \begin{tabular}{cccccc}
         \toprule
         Dataset & Statistics & Train & Dev & Test & Total \\
         \midrule
         \multirow{4}{*}{\alchemy} & \#Interaction & $3657$ & $245$ & $899$ & $4801$ \\
         & \#Instruction & $18285$ & $1225$ & $4495$ & $24005$ \\
         & Avg.inst/inte & $5$ & $5$ & $5$ & $5$ \\
         & Avg.word/inst & - & - & - & $8.0$ \\
         \midrule
         \multirow{4}{*}{\scene} & \#Interaction & $3352$ & $198$ & $1035$ & $4585$ \\
         & \#Instruction & $16760$ & $990$ & $5175$ & $22925$ \\
         & Avg.inst/inte & $5$ & $5$ & $5$ & $5$ \\
         & Avg.word/inst & - & - & - & $10.5$ \\
         \midrule
         \multirow{4}{*}{\tangrams} & \#Interaction & $4189$ & $199$ & $800$ & $5188$ \\
         & \#Instruction & $20945$ & $995$ & $4000$ & $25940$ \\
         & Avg.inst/inte & $5$ & $5$ & $5$ & $5$ \\
         & Avg.word/inst & - & - & - & $5.4$ \\
         \midrule
         \multirow{4}{*}{\propara} & \#Paragraph & $391$ & $43$ & $54$ & $488$ \\
         & \#Sentence & $2639$ & $290$ & $373$ & $3302$ \\
         & Avg.sent/para & $6.7$ & $6.7$ & $6.9$ & $6.8$ \\
         & Avg.word/para & $61.1$ & $57.8$ & $67.0$ & $61.4$ \\
         \midrule
         \multirow{4}{*}{\recipes} & \#Paragraph & $693$ & $86$ & $87$ & $866$ \\
         & \#Sentence & $6101$ & $766$ & $781$ & $7648$ \\
         & Avg.sent/para & $8.8$ & $8.9$ & $9.0$ & $8.8$ \\
         & Avg.word/para & $93.1$ & $89.1$ & $93.9$ & $92.8$ \\
         \bottomrule

    \end{tabular}
    \caption{Data statistics for \alchemy, \scene, \tangrams, \propara, \recipes.}
    \label{tab:statistics}
\end{table*}

\section{Downstream Performance \textit{w.r.t} Overlap Ratio}
\label{dp-wrt-or}
Table \ref{tab:overlap} shows the downstream performance on the validation sets with respect to the overlap ratio in the pre-training corpus. 

\begin{table*}[]
    \centering
    \begin{tabular}{ccccccc}
         \toprule
         \multirow{2}{*}{\textbf{Datasets}} & \multicolumn{6}{c}{\textbf{Overlap Ratio (\%)}} \\
         \cmidrule(lr){2-7}
         & $0$\% & $20$\% & $40$\% & $60$\% & $80$\% & $100$\% \\
         \midrule
         \alchemy & $81.0$ & $83.2$ & $81.2$ & $80.8$ & $82.0$ & $83.3$ \\ 
         \scene & $65.5$ & $63.1$ & $59.6$ & $62.6$ & $63.1$ & $64.6$ \\ 
         \tangrams & $62.8$ & $63.3$ & $62.8$ & $64.3$ & $62.8$ & $63.3$ \\ 
         \propara & $72.5$ & $70.7$ & $72.7$ & $70.0$ & $70.8$ & $72.8$ \\ 
         \recipes & 61.7 & 59.5 & 59.3 & 62.4 & 60.8 & 59.8 \\
         \bottomrule
    \end{tabular}
    \caption{Downstream Performances on the validation sets of five datasets with respect to the overlap ratio. For \alchemy, \scene and \tangrams, we report the 5utts denotation accuracy, while we report the F1 score on \propara and \recipes.}
    \label{tab:overlap}
\end{table*}

\section{Pre-training Improvement Analysis}
\label{sec:improvement-appendix}
The main types of the improvements by the execution-guided pre-training on the five tasks are shown in Table \ref{tab:pretrain-improvement-scone-appendix} and Table \ref{tab:pretrain-improvement-propara-recipes-appendix}.

\begin{table*}[t]
    \centering
    \small
    \scalebox{0.9}{
    \begin{tabular}{p{2cm}p{5.1cm}p{9.5cm}}
        \toprule
        \textbf{Type (Percent)} & \textbf{Example} & \textbf{Environment (Goal State) Comparison} \\
        \midrule
        \multicolumn{3}{c}{\textbf{\alchemy}} \\
        \midrule
        \multirow{2}{2cm}{Operation Correctness ($68.3$\%)} & \textbf{Environment (Initial State)} : 1:p | 2:r | 3:o & \multirow{2}{9.5cm}{The difference lies in the third beaker, $o$ (\textit{w.o.} pre-training) versus $\_$ (\textit{w.} pre-training). Without pre-training, the model does not correctly understand the semantics of ``add'', i.e., removing the liquid from the third beaker.} \\
        & \textbf{Instruction}: Empty out the first beaker, add the orange chemical to the red. & \\
        \midrule
        \multirow{2}{2cm}{Instruction Completeness ($18.3$\%)} & \textbf{Environment (Initial State)}: 1:o | 2:rr & \multirow{2}{9.5cm}{The difference lies in the first beaker, $orr$ (\textit{w.o.} pre-training) versus $or$ (\textit{w.} pre-training). As observed, without pre-training, the model seems to ignore the instruction ``throw out one unit of the second beaker''.} \\
         & \textbf{Instruction}: Throw out one unit of the second beaker, pour the second beaker into first one. & \\
        \midrule
        \multirow{2}{2.1cm}{Grounding Correctness ($8.5$\%)} & \textbf{Environment (Initial State)}: 1:y | 2:yyy | 3:yy & 
        \multirow{2}{9.5cm}{The whole states are 1:y | 2:yyy | 3:y (\textit{w.o.} pre-training) versus 1:y | 2:yy | 3:yy (\textit{w.} pre-training). Without pre-training, the model does not find the correct beaker according to the instructions.} \\
         & \textbf{Instruction}: Pour out one part of the second yellow beaker. & \\
         \midrule
        \multicolumn{3}{c}{\textbf{\scene}} \\
        \midrule
        \multirow{2}{2cm}{Operation Correctness ($22.7$\%)} & \textbf{Environment (Initial State)}: 1:og | 2:oo | 3:\_\_ | 4:\_\_ & \multirow{2}{9.5cm}{The difference lies in the second position, $\_\_$ (\textit{w.o.} pre-training) versus $oo$ (\textit{w.} pre-training). Without pre-training, the $oo$ disappears for no reason, which indicates that the ``move'' operation cannot be performed correctly. } \\
        & \textbf{Instruction}: The man in an orange shirt and green hat moves to the right end. & \\
        \midrule
        \multirow{2}{2cm}{Instruction Completeness ($15.2$\%)} & \textbf{Environment (Initial State)}: 1:\_\_ | 2:bo | 3:\_\_ | 4:\_\_ & \multirow{2}{9.5cm}{The difference lies in the second position and the third position, $b\_$, $yo$ (\textit{w.o.} pre-training) versus $bo$, $y\_$ (\textit{w.} pre-training). The model requires to swap hats twice, but without pre-training, only once performed, which indicates that one of the \textsc{Trade-hats} operations is ignored. } \\
        & \textbf{Instruction}: A person in a yellow shirt enters from the right, the person in yellow takes the hat from the person in blue, the person in blue retrieves the hat from the person in yellow. & \\
        \midrule
        \multirow{2}{2.1cm}{Grounding Correctness ($62.1$\%)} & \textbf{Environment (Initial State)}: 1:rg | 2:\_\_ | 3:gr | 4:\_\_ & \multirow{2}{9.5cm}{The whole states are 1:rg | 2:\_\_ | 3:gr | 4:o\_ (\textit{w.o.} pre-training) versus 1:rg | 2:o\_ | 3:gr | 4:\_\_ (\textit{w.} pre-training). Without pre-training, the model does not find the correct position according to the instructions. } \\
        & \textbf{Instruction}: A man in an orange shirt appears to the right of the man in a red shirt and green hat. & \\
        \midrule
        \multicolumn{3}{c}{\textbf{\tangrams}} \\
        \midrule
        \multirow{2}{2cm}{Instruction Completeness ($52.7$\%)} & \textbf{Environment (Initial State)}: 1:A | 2:C | 3:D & \multirow{2}{9.5cm}{The state is 1:C (\textit{w.o.} pre-training) versus 1:A (\textit{w.} pre-training). Without pre-training, the model does not correctly understand the semantics of ``swap'', i.e. change the positions of two figures. } \\
        & \textbf{Instruction}: Delete the 3rd figure, swap the two figures, delete the 1st figure. & \\
        \midrule
        \multirow{2}{2.1cm}{Grounding Correctness ($47.3$\%)} & \textbf{Environment (Initial State)}: 1:A | 2:E | 3:B | 4:C | 5:D & \multirow{2}{9.5cm}{The whole states is 1:A | 2:E | 3:D (\textit{w.o.} pre-training) versus 1:A | 2:E | 3:C (\textit{w.} pre-training). After performing the first instruction, the positions of each item have changed. When performing the second instruction, without pre-training, the model does not find the correct positions. } \\
        & \textbf{Instruction}: Delete the 3rd figure, delete the 4th figure. & \\
        \bottomrule
    \end{tabular}
    }
    \caption{The main types of the improvements by the execution-guided pre-training in the validation set of \alchemy, \scene, and \tangrams. }
    \label{tab:pretrain-improvement-scone-appendix}
\end{table*}

\begin{table*}[t]
    \centering
    \small
    \scalebox{0.9}{
    \begin{tabular}{p{2cm}p{5.1cm}p{9.5cm}}
        \toprule
        \textbf{Type (Percent)} & \textbf{Example} & \textbf{Environment (Goal State) Comparison} \\
        \midrule
        \multicolumn{3}{c}{\textbf{\propara}} \\
        \midrule
        \multirow{2}{2cm}{Operation Correctness ($12.0$\%)} & \textbf{Environment (Initial State)}: ent: algae | plankton | sediment; loc: ? | ? | seafloor & \multirow{2}{9.5cm}{The whole predicted locations contains 2 items (\textit{w.o.} pre-training) versus 3 items (\textit{w.} pre-training). Without pre-training, the amount of the predicted locations is inconsistent with the entities, namely, the instructions cannot be performed correctly. } \\
        & \textbf{Instruction}: Algae and plankton die. The dead algae and plankton end up part of sediment on a seafloor. & \\
        \midrule
        \multirow{2}{2cm}{Instruction Completeness ($28.0$\%)} & \textbf{Environment (Initial State)}: ent: bacteria | enzymes; loc: ground | bacterium & \multirow{2}{9.5cm}{The predicted location of enzymes is $bacterium$ (\textit{w.o.} pre-training) versus $plant\ material$ (\textit{w.} pre-training). The most potential reason is that the second instruction is ignored by the model without pre-training. } \\
        & \textbf{Instruction}: Bacteria from the ground migrate to the plant material. Bacteria release enzymes onto the plant material. & \\
        \midrule
        \multirow{2}{2cm}{Grounding Correctness ($60.0$\%)} & \textbf{Environment (Initial State)}: ent: silk | web; loc: - | - & \multirow{2}{9.5cm}{The predicted location of silk is $spider$ (\textit{w.o.} pre-training) versus $abdomen$ (\textit{w.} pre-training). Without pre-training, the model does not find the most suitable location of the entity based on the instructions. } \\
        & \textbf{Instruction}: The spider picks a suitable place. The spider produces sticky silk from its abdomen. & \\
        \midrule
        \multicolumn{3}{c}{\textbf{\recipes}} \\
        \midrule
        \multirow{2}{2cm}{Operation Correctness ($9.1$\%)} & \textbf{Environment (Initial State)}: ent: scallion | cloves garlic | canola oil states: - | - | - & \multirow{2}{9.5cm}{The whole predicted locations contain 4 items (\textit{w.o.} pre-training) versus 3 items (\textit{w.} pre-training). Without pre-training, the amount of the predicted locations is inconsistent with the entities, namely, the instructions does not be performed correctly. } \\
        & \textbf{Instruction}: First to go in the wok was the oil, scallions , and garlic .heat these ingredients until the garlic starts to turn brown. & \\
        \midrule
        \multirow{2}{2cm}{Instruction Completeness ($21.8$\%)} & \textbf{Environment (Initial State)}: ent: green pepper | tomato | green onion states: - | - | - & \multirow{2}{9.5cm}{All of the predicted locations of the three entities are $?$ (\textit{w.o.} pre-training) versus $foil$ (\textit{w.} pre-training). The most potential reason is that the second instruction is ignored without pre-training.} \\
        & \textbf{Instruction}: Cut tin foil into 12x16 inch rectangle. Place green pepper, tomato and green onion on lower half of foil sheet. & \\
        \midrule
        \multirow{2}{2cm}{Grounding Correctness ($69.1$\%)} & \textbf{Environment (Initial State)}: ent: salt | olive oil; loc: - | - & \multirow{2}{9.5cm}{The predicted location of olive oil is $bowl$ (\textit{w.o.} pre-training) versus $pan$ (\textit{w.} pre-training), which indicates that the model does not find the most suitable location of the entity based on the instructions without pre-training. } \\
        & \textbf{Instruction}: Lightly grease large bowl and two loaf pans with olive oil. & \\
        \bottomrule
    \end{tabular}
    }
    \caption{The main types of the improvements by the execution-guided pre-training in the validation set of \propara and \recipes. }
    \label{tab:pretrain-improvement-propara-recipes-appendix}
\end{table*}

\section{Example Program, Initial State and Goal State of Each Domain}
Table \ref{tab:program_example} shows the examples of each domain, including the initial environment state, the program, and the corresponding goal environment state.

\begin{table*}[t]
    \centering
    \small
    \begin{tabular}{llp{4cm}p{4.7cm}}
         \toprule
         Domain & Program & Initial State \& Goal State & Description \\
         \midrule
         \alchemy & \textsc{Pour} (\textsc{Beaker} (1), \textsc{Beaker} (2, g) ) & \begin{tabular}{p{3.5cm}}
              1:rr | 2:gg | 3:g | 4:ooo \\
              \midrule
              1:\_ | 2:gg | 3:grr | 4:ooo \\ 
         \end{tabular}   
         & Pour the liquid from the first beaker into the second green beaker. \\
         \midrule
         \scene & \textsc{Person} (2, r); \textsc{Hat} (2, y) & \begin{tabular}{p{3.5cm}}
              1:\_\_ | 2:\_\_ | 3:\_\_ | 4:\_\_ | 5:ob \\
              \midrule
              1:\_\_ | 2:ry | 3:\_\_ | 4:\_\_ | 5:ob \\ 
         \end{tabular}   
         & A person with a red shirt and a yellow hat appears on the second position. \\
         \midrule
         \tangrams & \textsc{Remove} (2); \textsc{Insert} (4, B) & \begin{tabular}{p{3.5cm}}
              1:A | 2:B | 3:C | 4:D | 5:E \\
              \midrule
              1:A | 2:C | 3:D | 4:B | 5:E \\ 
         \end{tabular}   
         & Remove the second figure, and add it back into the fourth position. \\
         \midrule
         \propara & \textsc{Move} (bacteria, cell, bladder) & \begin{tabular}{p{3.5cm}}
              ent : bacteria | sickness \\ 
              loc : cell | - \\
              \midrule
              ent : bacteria | sickness \\ 
              loc : bladder | - \\ 
         \end{tabular}   
         & Move bacteria from cell to bladder. \\
         \midrule
         \recipes & \textsc{Create} (beef, oven) & \begin{tabular}{p{3.5cm}}
              ent : beef | pepper \\
              loc : - | - \\
              \midrule
              ent : beef | pepper \\
              loc : oven | - \\ 
         \end{tabular}   
         & Beef appears in the oven. \\
         \bottomrule
         
    \end{tabular}
    \caption{Example programs, initial states, and goal states for each domain.}
    \label{tab:program_example}
\end{table*}

\section{Case Study}
\begin{figure}[t]
    \centering
    \includegraphics[width=1.0\columnwidth]{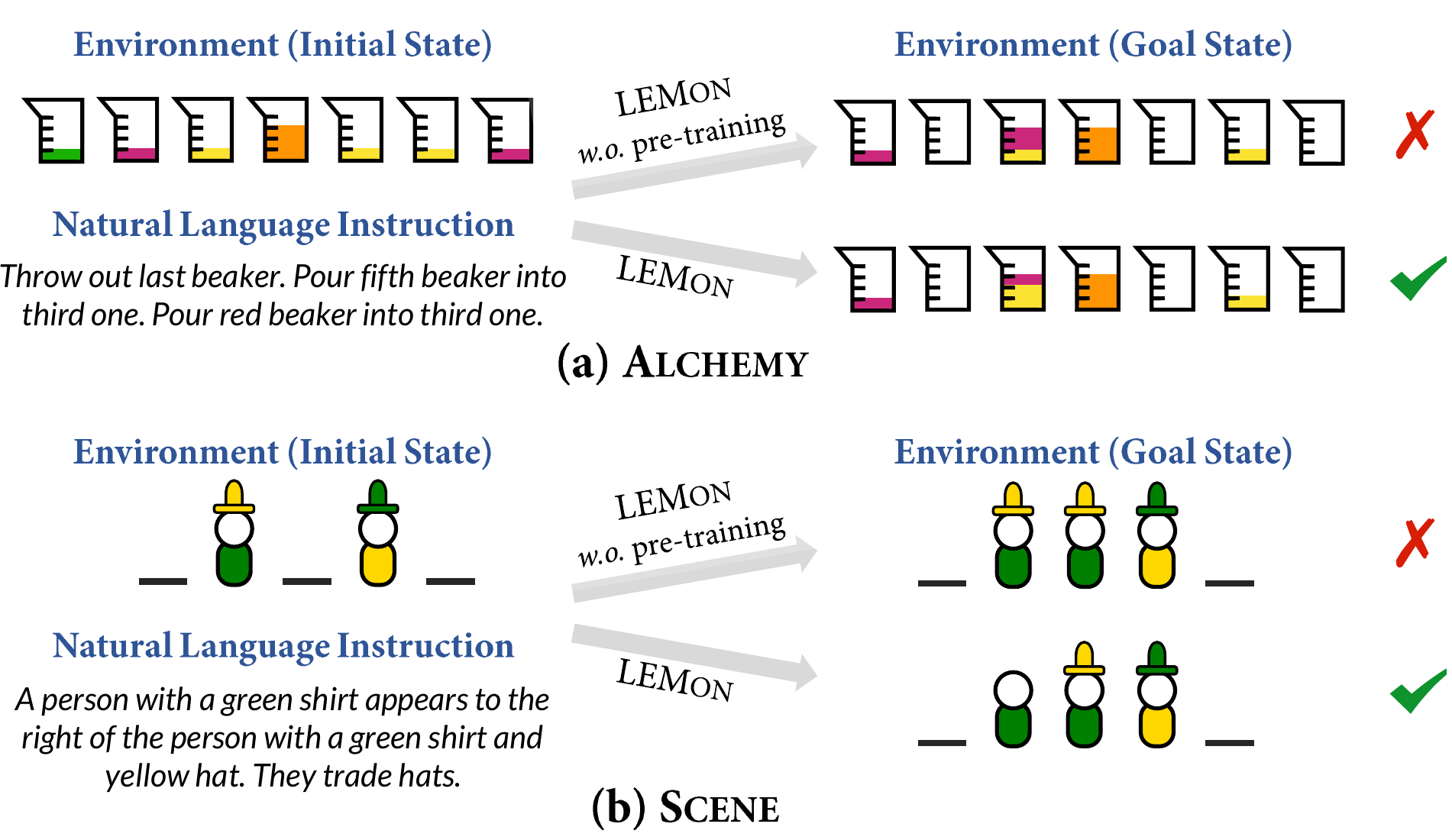}
    \caption{Two cases of \ours and \ours \textit{w.o.} pre-training in \alchemy and \scene. The predictions of \ours are more consistent with the semantics of the natural language instruction.}
    \label{fig:case_study}
\end{figure}

Figure \ref{fig:case_study} shows two cases in \alchemy and \scene, providing a more intuitive view of the role played by the execution-guided pre-training in \ours. 
We display the initial environment states, the natural language instructions, and the goal environment states predicted with\,/\,without applying the execution-guided pre-training strategy, respectively.
In the first case (a), when pouring \textcolor{cyellow}{{yellow}} liquid from the $5$-th beaker into the $3$-th beaker, the latter receives \textcolor{cred}{{red}} liquid, which is clearly an inconsistent change.
However, with pre-training, \ours can predict the correct goal environment state via deeply understanding the actions conveyed by natural language.
Similarly, in the second case (b), when swapping the hats in the last step, the model does not understand the \textsc{Trade-hat} action correctly, while it can be well understood to generate the goal state after pre-training. 
The above two cases indicate that the execution-guided pre-training strategy is able to inject prior knowledge of environments into \ours and benefit the downstream tasks.

\end{document}